\documentclass[acmsmall]{acmart}
\pdfoutput=1 
\AtBeginDocument{%
  \providecommand\BibTeX{{%
    \normalfont B\kern-0.5em{\scshape i\kern-0.25em b}\kern-0.8em\TeX}}}

\acmVolume{00}
\acmNumber{1}
\acmArticle{111}
\acmMonth{8}
\usepackage{float}
\usepackage{tikz}
\usepackage{graphicx}
\usepackage{multirow}
\usepackage{rotating}
\usepackage{subcaption}
\usepackage{caption}
\usepackage{todo}
\usepackage{array}
\usepackage[title]{appendix}

\usetikzlibrary{arrows}
\tikzset{
      treenode/.style = {align=center, inner sep=0pt, text centered,
        font=\sffamily},
      arn_n/.style = {treenode, circle,black, draw=black, text width=4em},
      arn_r/.style={treenode,circle,black,draw=red,dashed,text width=4em},
    level distance = 2cm,
    level 1/.style={sibling distance=3cm},
    level 2/.style={sibling distance=4cm},
    level 3/.style={sibling distance=2cm}
}
\usepackage{amssymb}
\usepackage{pgfplots}
\pgfplotsset{width=15cm,compat=1.8}
\usepackage{filecontents}

\begin{document}

\title{An empirical investigation of different classifiers, encoding and ensemble schemes for next event prediction using business process event logs}

\author{Bayu Adhi Tama}

\email{bayuat2802@postech.ac.kr}
\orcid{0000-0002-1821-6438}
\affiliation{%
  \institution{Department of Mechanical Engineering, Pohang University of Science and Technology (POSTECH)}
  \streetaddress{77 Cheongam-Ro, Nam-Gu}
  \city{Pohang}
  \state{Republic of Korea}
  \postcode{37673}
}

\author{Marco Comuzzi}
\authornote{Corresponding author: Marco Comuzzi (mcomuzzi@unist.ac.kr)}
\email{mcomuzzi@unist.ac.kr}
\affiliation{%
  \institution{School of Management Engineering, Ulsan National Institute of Science and Technology (UNIST)}
  \streetaddress{50, UNIST-gil}
  \city{Ulsan}
  \state{Republic of Korea}
  \postcode{44919}
}

\author{Jonghyeon Ko}
\email{whd1gus2@unist.ac.kr}
\affiliation{%
  \institution{School of Management Engineering, Ulsan National Institute of Science and Technology (UNIST)}
  \streetaddress{50, UNIST-gil}
  \city{Ulsan}
  \state{Republic of Korea}
  \postcode{44919}
}

\renewcommand{\shortauthors}{Tama, Comuzzi and Ko}

\begin{abstract}
There is a growing need for empirical benchmarks that support researchers and practitioners in selecting the best machine learning technique for given prediction tasks. In this paper, we consider the next event prediction task in business process predictive monitoring and we extend our previously published benchmark by studying the impact on the performance of different encoding windows and of using ensemble schemes. The choice of whether to use ensembles and which scheme to use often depends on the type of data and classification task.  While there is a general understanding that ensembles perform well in predictive monitoring of business processes, next event prediction is a task for which no other benchmarks involving ensembles are available. The proposed benchmark helps researchers to select a high performing individual classifier or ensemble scheme given the variability at the case level of the event log under consideration. Experimental results show that choosing an optimal number of events for feature encoding is challenging, resulting in the need to consider each event log individually when selecting an optimal value. Ensemble schemes improve  the performance of low performing classifiers in this task, such as SVM, whereas high performing classifiers, such as tree-based classifiers, are not better off when ensemble schemes are considered.
\end{abstract}

\setcopyright{acmcopyright}
\acmJournal{TIST}
\acmYear{2020} \acmVolume{1} \acmNumber{1} \acmArticle{1} \acmMonth{1} \acmPrice{15.00}



\maketitle

\section{Introduction}

Process mining is a discipline in business process management that aims at analyzing and extracting previously unknown and useful information about a business process derived from historical data about process execution \citep{li2016process, schonig2016framework}. A prominent topic in process mining is predictive monitoring of business processes~\citep{de2016general}. The objective of predictive monitoring is to use historical process data, normally stored in so-called \emph{event logs}, to predict some aspect of interest about currently running cases of a business process. Prediction may concern either aspects related with execution of process cases, such as predicting the next activities that will be executed in a case or the time at which future activities are likely to happen~\citep{polato2018time} or \emph{outcomes} of a process case, such as the satisfaction or violation of given service level objectives~\citep{leitner2009runtime} or logical constraints predicated, for instance, on the possible occurrence and order of activities~\citep{baier2015matching}. Information obtained from predictive monitoring is exploited to support pro-active decision making during business process execution, e.g., warning customers that their request may be honored later than planned or taking corrective actions if an undesired exception is likely to occur with high probability.

Process predictive monitoring models can be generated using classification or regression techniques~\citep{di2016clustering, maggi2014predictive, tax2017predictive, di2018genetic}. The type of technique chosen for predictive monitoring depends on the variable to be predicted, whether discrete or continuous. Classification and regression techniques deal with the prediction of discrete and continuous variables, respectively. For instance, the authors of ~\citep{tax2017predictive} propose a method based on Long Short-Term Memory (LSTM) neural networks to predict the remaining time of a  process instance, which can be viewed as solving a regression problem. Alternatively, the task of predicting the next event most likely to happen in a process case is a sequence classification problem that can be solved using classification techniques, such as support vector machines or decision trees~\citep{verenich2016minimizing}. Similarly, predicting process outcomes can also be seen as a classification problem, i.e., early time series classification~\citep{castellanos2006predictive,zeng2008event}.

An open issue in predictive monitoring is the one of choosing an appropriate and high performing machine learning technique in a given scenario~\citep{maggi2014predictive,marquez2017predictive,di2018predictive}. This is a challenging endeavour since every dataset, i.e., event log, is likely to be different, owing to the fact that the organisational context and the business process that generate event logs may vary dramatically in practice, spanning from patient management in a hospital, to IT incident management or manufacturing processes. Given the high variability of dataset characteristics in the context of process predictive monitoring, it is desirable to build a comparative study of various machine learning techniques for different predictive monitoring tasks, in order to support decision-makers in choosing the best machine learning model for the predictive monitoring task at hand. Quantitative benchmarks of different machine learning techniques for predicting the remaining time of process cases~\citep{verenich2019survey}, outcomes of a process case~\citep{teinemaa2019outcome} or next event prediction~\cite{tama2019empirical} have been recently published in the literature. 

In this paper, we present an extension of our previously published quantitative benchmark~\cite{tama2019empirical} for the next event prediction task. This extension addresses (i) the effect of different encoding, in particular, the size of the window chosen to generate samples from a trace of events, on the performance of the classifiers, and (ii) the opportunity of using ensemble classifiers for next event prediction and, in particular, whether it is worth to choose an ensemble for this prediction task.

Regarding the encoding of features, the previously published benchmark~\cite{tama2019empirical} and other published work, e.g.~\cite{marquez2017run,tax2017predictive},  have made an arbitrary choice of generating features from attributes of a fixed number of events, i.e., a \emph{window}, preceding the one to predict. In this paper, we benchmark different sizes of such a window, showing that choosing an optimal window size is a challenging endeavour and that the optimal choice may differ for each event log considered.   

Regarding the assessment of whether it is worth to use ensemble learners in the particular task of next event prediction, in this paper, we focus specifically on the issue of comparing different ensemble schemes, and on the related issue of choosing the best base classifier to use in an ensemble scheme. Ensemble classifiers, in fact, have been shown to perform well in process predictive monitoring. In the context of outcome-based predictive monitoring, \citep{teinemaa2019outcome} reports eXtreme Gradient Boosting (XGBoost) using decision tree as a base classifier as the best overall performer in more than 50\% of the considered datasets. The performance, however, is evaluated against a limited number of alternative base classifiers.

Despite ensembles being adopted effectively by previous research, the selection of available individual classifiers in an ensemble may require prior knowledge about a dataset. Moreover, researchers are often accustomed to particular individual classification algorithms, which they tend to select as base classifiers. Therefore,  somehow individual base classifiers in ensembles are often picked arbitrarily, without taking into account other classifiers outside the researcher's proficiency. In addition, when proposing a new classifier ensemble, researchers tend to include only a small number of well known individual classifiers, such as decision trees and neural networks, without exploring extensively the whole scope of classification techniques available.

Hence, a comparative analysis among classifier ensembles and individual classifiers spanning from different families, i.e., trees, rule-based, Bayes, and neural-based classifiers, on different types of event logs is currently lacking. In this paper, we consider 12 base classifiers and 5 ensemble schemes. We also consider 6 different real world event logs commonly used for benchmarking in the process mining community, which are produced by different types of business processes.

In line with the 'no free lunch theorem', while some ensemble schemes may perform better with particular event logs, the best performers will vary over different event logs~\citep{wolpert1997no}. In evaluating the performance of different ensemble schemes across different event logs, our objective is twofold. First, we aim at identifying the best performing base classifier for a given ensemble scheme across all event logs. This is important for allowing decision-makers to make a more informed choice when adopting ensemble learning in predictive monitoring, allowing them to choose an ensemble scheme likely to perform well across all possible scenarios. Second, we aim at studying the performance of ensemble schemes across different groups of events logs with different characteristics. In particular, we consider event logs characterized by small and large variability at the case level. The results show that ensemble schemes improve the performance of classifiers performing low when considered as individuals, while high performing classifiers, particularly in terms of low variability event logs, tend to be stable, i.e., their performance does not improve particularly when considered in ensemble schemes. 

To summarise, this extended version of the benchmark in~\cite{tama2019empirical} for next event prediction addresses the following research questions:

\begin{itemize}
    \item RQ1: What is the impact of the encoding of features and, in particular, the size of the window of events considered to make a prediction, on the performance of a classifier? 
    \item RQ2: Do ensemble schemes improve the performance of base classifiers and, if so, is there a best choice of the combination of ensemble scheme and base classifier? 
\end{itemize}

The variability of event logs, in terms of frequency distribution of trace variants, is also considered as a variable in the proposed benchmark. Therefore, for each question we investigate to what extent the findings change with different levels of event log variability.

The remainder of the paper is organized as follows. Section~\ref{rev} discusses related work. The configuration of the proposed comparative analysis is detailed in Section~\ref{expconf}, while Section~\ref{expres} reports and discusses the experimental results. Finally, concluding remarks are presented in Section~\ref{conc}.

\section{Related Work}
\label{rev}
Table~\ref{tab4} summarizes chronologically the existing techniques for predictive monitoring of business processes that use classification techniques using the following criteria: (i) whether the event logs used are private or publicly available, (ii) the classification method used, (iii) the main performance measure considered, (iv) the type of prediction objective, i.e., whether next event or outcome prediction, and (v) whether a significance test to compare the performance of different classifiers has been used.  Decision tree appears as popular choice as a classification algorithm (7 publications), followed by random forest (6 publications), and support vector machine (4 publications). Few works have considered classifier ensembles, i.e., random forests and XGBoost, in their studies~\citep{leontjeva2015complex, teinemaa2016predictive,klinkmuller2018towards,santoso2018specification,teinemaa2019outcome, di2018genetic}. This indicates that classifier ensembles are still unexplored in the published works. Moreover, only one study~\citep{teinemaa2019outcome} considers a statistical test to compare the performance of different ensemble schemes.

\begin{table}
\caption{\label{tab4} Outline of predictive monitoring using classification algorithms}
\centering
\resizebox{1\textwidth}{!}{
\begin{tabular}{llp{1.8cm}p{2.5cm}p{2cm}p{2.4cm}p{2cm}}
\hline
Study&Year&dataset&Method&Performance measure& Prediction objective&Significance test\\
\hline\hline
\cite{kang2012periodic}&2012&Private&Support vector machine&Error rate&Process outcome&No\\
\cite{leitner2013data}&2013&Private&Decision tree \& neural network&Precision&Process outcome&No\\
\cite{cabanillas2014predictive}&2014&Private&Support vector machine&F-score&Next event&No\\
\cite{maggi2014predictive}&2014&BPIC2011&Decision tree&Precision&Process outcome&No\\
\cite{breuker2014designing}&2014&BPIC2012&Expectation-maximization&Accuracy&Process outcome&No\\
\cite{conforti2015recommendation}&2015&Private&Decision tree&N/A&Process outcome&No\\
\cite{leontjeva2015complex}&2015&BPIC2011&Hidden Markov Model and Random forest&AUC&Process outcome&No\\
\cite{unuvar2016leveraging}&2016&Marketing campaign&Decision tree&Accuracy&Next event&No\\
\cite{di2016clustering}&2016&BPIC2011&Decision tree&Accuracy&Next event&No\\
\cite{teinemaa2016predictive}&2016&Private&Random forest and logistic regression&F-score&Next event&No\\
\cite{verenich2016minimizing}&2016&Private&Support vector machine&AUC&Process outcome&No\\
\cite{marquez2017run}&2017&BPIC2013&Evolutionary computing&F-score&&No\\
\cite{mehdiyev2017multi}&2017&BPIC2012, BPIC2013, and Helpdesk&Multi-stage deep learning&Accuracy, Precision, and Recall&Next event&No\\
\cite{evermann2017predicting}&2017&BPIC2012 and BPIC2013&Long Short Term Memory Neural Network&Precision&Process outcome&No\\
\cite{klinkmuller2018towards}&2018&Synthetic dataset&Random forest&Precision and Recall&Next event&No\\
\cite{di2018genetic}&2018&BPIC2011 and BPIC2015&Decision tree and random forest&Accuracy&Process outcome&No\\
\cite{santoso2018specification}&2018&BPIC2013&Decision tree and Random forest&Accuracy and AUC&Next event and process outcome&No\\
\cite{teinemaa2019outcome}&2019&BPIC2011, BPIC2012, BPIC2015, BPIC2017, Production, Insurance, Sepsis, Hospital Billing, and Traffic Fines&random forest, XGBoost, logistic regression, and support vector machine&AUC and F-score&Process outcome&Yes\\ 
\hline
\end{tabular}}
\end{table}

As far as reviews and benchmarks of different predictive monitoring techniques are concerned, qualitative reviews of research works have been proposed by~\citep{di2018predictive} and \citep{marquez2017predictive}. The review in \citep{di2018predictive}  provides a qualitative value-driven analysis of different predictive process monitoring techniques to support decision-makers in choosing the best predictive technique for a given task. The criteria considered in the classification framework are the type of prediction task considered, the input type, e.g., whether an event log is provided with additional data and/or contextual information, the family of algorithms and available tool support.
The review presented in~\citep{marquez2017predictive} considers standard criteria for classifying predictive monitoring approaches in the literature, such as the prediction task or the type of technique used. Additionally, it characterizes approaches in the literature according to their process awareness, i.e., whether or not an approach harnesses an explicit representation of process models. 

A genetic algorithm-based method for hyperparameter optimization in predictive monitoring is presented in~\citep{di2018genetic}. In this work,  decision tree and random forest are considered in the task of predicting process outcomes, formulated as satisfaction of linear temporal logic constraints.  In the paper, the authors argue that there is no single algorithm, under the default learning parameters, that constantly performs best across all process event logs, therefore calling for the development of quantitative benchmarks for business process predictive monitoring. 
Regarding the prediction of time aspects, the work presented in~\citep{verenich2019survey} benchmarks two regression algorithms based on XGBoost~\citep{chen2016xgboost} and LSTM neural networks~\citep{hochreiter1997long} for predicting the remaining time of process cases. The results indicate that in 14 of 17 datasets, LSTM had been the best-performing regressor. 
Regarding the prediction of process outcomes, the work in~\citep{teinemaa2019outcome} develops a benchmark of 4 classification algorithms on several publicly available event logs. The benchmark yields the XGBoost classifier as the best performer in terms of AUC metric in 15 out of 24 datasets. Regarding next event prediction, our previously published benchmark~\cite{tama2019empirical} provides an empirical comparison of the performance of 20 different classifiers, including 5 ensemble learners that use decision tree as the only base classifier. 

Overall, the existing quantitative benchmarks in business process predictive monitoring (\citep{verenich2019survey,teinemaa2019outcome,tama2019empirical} consider only a limited number of ensemble learners with default base classifiers, never attempting to study whether and how the choice of ensemble learner and base classifier may impact the performance of the model. 

In the general field of machine learning, it is widely recognised that choosing the best base classifier when designing an ensemble scheme is a challenging task. As a result, for the sake of generalization, there are no context rationales to prefer one ensemble to another~\citep{Duda:2000:PC:954544}. Even though there exists a common understanding among machine learning researchers that the performance of ensembles often surpasses the one of an individual classifier, this is not promised for all possible datasets~\citep{kuncheva2014combining}. Therefore, given a specific type of dataset, there exists a research gap related to choosing the best base classifier for particular ensemble schemes.         
Owing to the availability of public datasets, credit scoring is among the domains in which the effectiveness of ensemble learners has been investigated extensively. The works~\citep{wang2011comparative, marques2012exploring, abellan2014improving, abellan2017comparative} compare the performance of base classifiers and ensemble learners on multiple credit scoring datasets. Despite the fact that ensemble learners bring significant performance improvements over single classifiers, only a small number of classifiers are included in these studies. 
A novel contribution evaluating classifier ensembles for intrusion detection systems is presented in~\citep{tama2017extensive} and~\citep{tama2017detailed}. The benchmarks, however, consider only a few particular datasets, i.e., wireless and wired networks, because benchmark datasets in this domain are generally not publicly available. In addition, the performance of several tree-based classifiers for disease prediction, e.g. diabetes, is assessed either as a single classifier or in ensemble~\citep{tama2017tree}. This research, however, is restricted to a limited number of families of classifiers.

\section{Problem definition and experimental benchmark configuration}
\label{expconf}
This section describes the configuration of the proposed comparative analysis, covering the problem definition, materials (i.e., event log datasets), a description of the considered base classifiers and ensemble techniques, and a brief explanation of the significance tests adopted to investigate the performance differences among classifiers.    

\subsection{Problem definition}

\begin{table}[t]
\caption{\label{tab1}Process log example for the next event prediction}
\centering
\begin{tabular}{lll}
\hline
case\_id&event\_type&timestamp\\
\hline\hline
173688&activity A&10/01/2011 19:45\\
173688&activity B&10/01/2011 20:17\\
173688&activity C&10/13/2011 18:37\\
&&\\
173691&activity A&10/01/2011 19:43\\
173691&activity B&10/01/2011 22:36\\
173691&activity C&10/10/2011 19:30\\
173691&activity C&10/10/2011 22:17\\
$\cdots$&$\cdots$&$\cdots$\\
\hline
\end{tabular}
\end{table}

An event is a tuple $e=\left( c, t, a, (d_1, v_1), \ldots, (d_I, V_I) \right)$, where $c$ is the id of the process case to which $e$ belongs, $t$ is the timestamp at which $e$ has been recorded, $a$ is the event type, e.g., the activity that was executed, and $(d_i, v_i)$, for $i=1,\ldots, I$ are a set of $I$ domain specific attribute-value pairs associated with $e$. We refer to $\mathcal{A}$ as the universe of event types. We use the notation $\#_x(e_i)$, with $x \in \left\{ c,t,a,d_1,\ldots, d_I \right\}$, to refer to the value assumed by a particular attribute $x$ in an event $e_i$.

A trace $\sigma$ is the sequence of events executed in a particular process case, i.e., $\sigma = \left[ e_1, \ldots, e_n, \ldots, e_N  \right]$, with $\#_t(e_{n+1}) > \#_t(e_n) \forall n \in[1,N-1]$ and $\#_c(e_i) = \#_c(e_j), \forall i,j \in [1,N]$. Table~\ref{tab1} shows an example of  events for 2 traces with case id 173688 and 173691 in an event log where only the timestamp attribute is shown.
Let us refer to $\mathcal{E}$ and $\mathcal{S}$ as the universe of events and sequences of events (i.e., including traces), respectively. 

A window function $W:\mathcal{S} \times \mathcal{E} \times \mathbb{N} \longrightarrow \mathcal{S}$ maps a trace $\sigma$, an event $e_i \in \sigma$ onto a window of $l$ events preceding $e_i$ in $\sigma$:

\[
w(\sigma, e_i, l) = \left\{
  \begin{array}{lr}
    \left[ e_{i-l}, \ldots, e_{i-1}\right] \subseteq \sigma & \text{if } i -l \geq 1\\
    \bot & \text{otherwise}
  \end{array}
\right.
\]

We refer to $l$ as the \emph{size} of the window $w(\sigma, e_i, l)$. 

A window generation relation $g: \mathcal{S} \times \mathbb{N} \longrightarrow 2^{\mathcal{S}}$ maps a trace $\sigma$ into the set of all possible windows of size $l$ generated from it:

\[
g(\sigma, l) = \left\{ w(\sigma, e_i, l) \neq \bot, \forall e_i \in \sigma \right\}
\]

A window encoder function $e: \mathcal{S} \longrightarrow \mathcal{X}_1, \ldots, \mathcal{X}_f, \ldots,  \mathcal{X}_F$ maps the sequence of events in a window onto a vector of $F$ features $\mathcal{X}_f$, with $f \in [1,F]$. Finally, a window labelling function $y:\mathcal{S} \longrightarrow \mathcal{A}$ maps a sequence of events in a window onto its event type:

\[
y(w(\sigma, e_i, l)) = \#_a(e_i)
\]

The problem of next event prediction is to learn a classifier function $cls: \mathcal{X}_1, \ldots, \mathcal{X}_f, \ldots,  \mathcal{X}_F \longrightarrow \mathcal{A}$ mapping a feature vector onto its label.  With an abuse of notation, in the remainder we use $cls(w)$ to indicate the output of the classifier applied to the features generated from a window of events $w(\cdot)$.

Note that this problem differs from the one of prediction of outcomes in business process predictive monitoring. Predictions of outcomes is normally treated as an instance of early time series classification~\cite{teinemaa2019outcome}, where the aim is to predict as soon as possible in a case what the outcome of that case will be. As such, features in prediction of outcomes are derived from prefixes of cases, which start from the first event registered in a case. Next event prediction, as formalised above,  is an instance of sequence classification, in which a long sequence, i.e., a trace, can be broken down into a set of small consecutive labeled sequences, i.e., using the window-based encoding described above, to create the samples for the classification task~\cite{sun2001sequence}. Next event prediction may also be treated as an instance of early time series classification, by considering the label of the next event in a case as the outcome. However, in this paper we decided to consider the window-based encoding because it is used extensively by previous research~\cite{tama2019empirical,marquez2017predictive}.

\subsection{Materials}
We consider 6 event logs. Besides being publicly available, event logs have been chosen because they are generated by different types of business processes and because they differ in terms of variability at the case level, e.g., the number and frequency of case trace variants.  In most cases, the predicted event type $a_i$ corresponds to an activity label, i.e., the execution of a particular instance of a business process. However, in some cases, an event type might represent different information that could still be utilized for process-aware analysis of event logs, such as the status of customer loan applications in a financial institution. In the event logs used in this study, when not explicitly defined, a feature \emph{activity} is available to be taken into account as event type. The 6 event logs are described next.

\begin{enumerate}
    \item Helpdesk\footnote{http://dx.doi.org/10.17632/39bp3vv62t.1}\\
    This log records events from a ticketing management system of the help desk of an Italian software company. The log has 9 event types (i.e., distinct activities), 3,804 process cases and 13,710 events.      
    
    \item BPIC 2012\footnote{http://dx.doi.org/10.4121/uuid:3926db30-f712-4394-aebc-75976070e91f}\\
    This event log has been made available by the Business Process Intelligence Challenge (BPIC) in 2012. It records events from the application procedure for financial products in a large financial institution. The log comprises 262,200 events in 13,087 cases. The event type considered is the combination of the attributes \texttt{concept:name} and \texttt{lifecycle:transition} in the log, which denotes the status of applications. This log includes 36 different event types.      
   
    \item BPIC 2013\footnote{http://dx.doi.org/10.4121/uuid:500573e6-accc-4b0c-9576-aa5468b10cee}\\
    This event log has been made available by the BPIC in 2013. It records the events of an incident and problem management system at a car manufacturer in Belgium. It comprises 65,533 events for 7,554 cases. The event type is an activity label obtained as the combination of the attributes \texttt{concept:name} and \texttt{lifecycle:transition}. This log has 13 different event types.           
    
    \item Sepsis\footnotemark \footnotetext{https://www.bupar.net/eventdataR.html}\\
    This log records events of sepsis cases treatment from a hospital as recorded by the hospital's enterprise resource planning (ERP) system. It has 15,214 events for 1,050 cases, and 16 different event types (distinct activities). 
    
    \item Road Traffic Fine Management\footnotemark[\value{footnote}]\\
    This event log records events from an information system managing road traffic fines for the local police of a city in Italy. It contains 34,724 events for 10,000 cases, with 11 different event types (distinct activities).
    
    \item Hospital Billing\footnotemark[\value{footnote}]\\
    This event log records events of the billing of medical services from the financial modules of a regional hospital's ERP system. The event log includes 49,951 events for 10,000 cases, with 16 different event types (distinct activities).
    
\end{enumerate}

We split the 6 considered event logs into two groups based on their variability at the case level (see Table~\ref{tab:logs}). In the \emph{low} variability group, a small number of case variants are needed to cover the large majority (80\%) of cases in the log, which means that the process generating the event log is often executed following a limited number of possible ways. Conversely, a much higher number of case variants is needed to cover 80\% of the cases in \emph{high} variability event logs. The variability of logs is numerically captured by the variability \emph{ratio}, defined as the ratio between the number of trace variants that cover 80\% of the cases and the total number of case variants. Note that there is at least one order of magnitude difference in the value of this ratio between low and high variability logs. Predictive monitoring in low variability event logs is normally an easier task, that is, models trained using this type of logs tend to be more stable and accurate~\citep{tama2019empirical}. 

\begin{table}
\centering
\caption{\label{tab:logs}A grouping of event log with respect to its variability level~\citep{tama2019empirical}}
\resizebox{1.05\textwidth}{!}{
\begin{tabular}{l|l|l|l|l|l|l}
\hline
Event log & Event types & Trace variants & \parbox{2.0cm}{Number of variants to 80\% cases} & Variability (ratio) &\parbox{2.0cm}{Mean trace length}&\parbox{2.0cm}{Median trace length}\\
\hline \hline
Helpdesk & 9 & 154 & 5 & Low (0.032)&3.60&3 \\
Hospital Billing & 16 & 288 & 3 & Low (0.01) &5.00&5 \\
Road Traffic & 11 & 44 & 2 & Low (0.045) &3.47&2 \\
\hline
Sepsis & 16 & 846 & 635 & High (0.75)  &14.49&13 \\
BPIC2013 & 13 & 2278 & 767 & High (0.33) &20.04&11 \\
BPIC2012 & 36 & 4366 & 1748 & High (0.40) &8.68&6 \\
\hline
\end{tabular}
}
\end{table}

Features are generated considering the event type of events in a window and the duration of a window. Similarly to~\cite{tama2019empirical,marquez2017run}, in fact, given a window $[e_{i-l}, \ldots, e_{i-1}]$ of size $l$, we consider as features the event types $\#_a(e_{i-l}), \ldots, \#_a(e_{i-1})$ and the duration of the considered window calculated as $\#_{t}(e_{i-1}) - \#_{t}(e_{i-l})$. As previously discussed, the label for a window  is the type of the next event, i.e.,  $\#_a(e_i)$. We consider 3 different window sizes $l=3,4,5$.

For example, Table~\ref{tab2} shows an example of encoding the event log of Table~\ref{tab1} for $l=3$, where the event type labels have been encoded into numerical attributes (Activity A into 3, B into 5, and C into 6). Note there is no optimal solution for choosing the window size $l$ for encoding and researchers, such as in~\citep{marquez2017run} and \citep{tax2017predictive}, tend to consider only a fixed window size in their experiments.

\begin{table}
\caption{\label{tab2}Final format of event log example shown in Table \ref{tab1} in for $l=3$}
\centering
\begin{tabular}{llllll}
\hline
case\_id&event\_1&event\_2&event\_3&duration&next\_event\\
\hline\hline
173688&3&5&6&17212&3\\
173691&3&5&6&12947&6\\
173691&5&6&6&12941&6\\
$\cdots$&$\cdots$&$\cdots$&$\cdots$&$\cdots$&$\cdots$\\
\hline
\end{tabular}
\end{table}

\subsection{Classification Techniques}

This section outlines the considered base classifiers and ensemble schemes. Besides a brief description, for each classifier we mention the implementation that we considered and the values that we considered in experiments for the main parameters. Each classifier, when not explicitly mentioned, runs using the default learning parameter settings (a list of hyperparameters used in this study is detailed in the Appendix~\ref{param}). 

\subsubsection{Individual Classifiers}

\begin{enumerate}
\item Decision Tree (DT)\\
We consider the $J48$ algorithm implementation of C4.5~\citep{quinlan2014c4}. Decision tree is a well known classification algorithm, where a tree is formed by a root and a number of nodes. Each node refers to a class label  and samples are assigned to nodes based on the impurity level of the class label distribution. In our experiment, tree-pruning is performed with confidence factor equal to 0.25.

\item  Credal Decision Tree (C-DT)\\
We consider the $JCDT$ implementation of the credal decision tree~\citep{abellan2003building}. This classifier, unlike $C4.5$, uses imprecise probabilities and uncertainty measures for assigning samples to nodes. In the experiment, tree-pruning is applied, the parameter used to fix the root node of the tree ($k$-th root variable) and the parameter used in the Imprecise Dirichlet Model ($S_{value}$) are both set to 1.

\item Random Tree (RT)\\
This classifier is a decision tree that uses $K$ randomly selected attributes at each node to build the tree, without pruning~\citep{breiman2001random}. In the experiment, we consider the implementation of this classifier in Weka. The parameter maximum depth of the tree is set to \emph{unlimited}, whereas the number of randomly chosen attributes is set to 0.  

\item Decision Stump (DS)\\
This is a 1-level decision tree, where the root is immediately connected to the leaves~\citep{iba1992induction}. It is commonly employed as a base classifier in boosting ensembles. In this study, we use the implementation of this classifier provided in Weka.

\item Naive Bayes (NB)\\
This classification technique takes into account the conditional probabilities of a categorical class variable defined by an independent predictor variables using the Bayes rule~\citep{john1995estimating}. It assumes independence of the predictor attributes. We consider the Java implementation of the classifier available in Weka. We consider a normal distribution instead of a kernel estimator since we deal with both categorical and numeric attributes.  

\item Support Vector Machine (SVM)\\
This classifier generates a set of \emph{hyperplanes} in a higher dimensional space used for classification and regression~\citep{cortes1995support}. As suggested by~\citep{hsu2003practical}, we use a LIBLINEAR~\citep{fan2008liblinear} implementation because it is faster than other implementations, such as LIBSVM~\citep{chang2011libsvm}, to achieve a classification model with comparable accuracy. In the experiment, we employ a $L2$-loss support vector classification (dual). The tolerance of the termination criterion $\epsilon$ is set to 0.01, the cost parameter $C$ to 1.0, and the maximum number of iterations $n$ to 1000.

\item $k$-Nearest Neighbor ($k$-NN)\\
The $k$-nearest neighbor classifier does not have an explicit training process. For a test sample, it calculates the $k$ samples from the training set that are nearest to the test sample. Next, the test sample is classified by choosing the majority class among the $k$ samples~\citep{altman1992introduction}. In our experiment, we use the $IBk$ implementation~\citep{aha1991instance} provided by Weka. The number of neighbors to use $k$ is set to 2, while the linear search using Euclidean distance is considered as the nearest neighbor search algorithm.  

\item RIPPER (JRip)\\
Repeated Incremental Pruning to Produce Error Reduction (RIPPER) has been originally proposed to improve the performance of the $IREP$ algorithm~\citep{cohen1995fast}. It generates a classification rule by  (i) splitting the samples randomly into two disjoint subsets, i.e., a growing set and a pruning set, and (ii) generating classification rules using the FOIL algorithm. Once a rule is generated, it is immediately pruned by repealing any final sequence of conditions. We consider the Java implementation of this classifier available in Weka, namely $JRip$.

\item OneR\\
OneR is a simple classification algorithm that yields one rule for each predictor in a dataset and finds the rule with minimum total error as its final rule. A rule for a predictor is obtained by creating a frequency table for each predictor variable against the target variable~\citep{holte1993very}. We consider the implementation provided by Weka, where the minimum bucket size used for discretizing numeric attributes is set to 6.   

\item Conjunctive Rule (CR)\\
This classifier generates rules in which the successor considers the distribution of the number of classes in the dataset, whereas the  predecessor is found by calculating the information gain of each predecessor and cutting off the resulting rules using reduced error pruning ($REP$). This procedure reduces the complexity of the final classifier, as a rule with small prediction would be pruned~\citep{friedman2001elements}. In the experiment, we consider the implementation provided in Weka, where pre-pruning for predecessors is performed.  

\item Bayesian Network (BN)\\
Bayesian network is a probabilistic graphical model that represents a set of attributes and their conditional dependencies \citep{pearl1985bayesian}. For a classification task, it learns the network structure and the probability tables defining conditional dependencies. We consider the $BayesNet$ implementation of this classifier in in Weka using the $simple$ estimator for searching the conditional probability tables. The method used for finding network structures is the $K2$ algorithm~\citep{cooper1992bayesian}.   

\item Decision Table and Naive Bayes Hybrid Classifier (DTNB)\\
The DTNB in a Bayes classifier that splits the attributes of a dataset into two disjoint subsets, one partition for decision tables and the other for naive Bayes. A forward selection search is used to assess the merit of subsets. At first, all attributes are modeled using decision tables, while at each assessment step, selected attributes are modeled by naive Bayes and the rest by decision tables \citep{hall2008combining}. We consider the implementation of this classifier provided by Weka. The search method used to find attribute combinations for the decision table is the $BackwardWithDelete$ algorithm, whilst the \emph{leave-one-out}  technique is used to evaluate the features.
   
\end{enumerate}

\subsubsection{Ensemble Schemes}
\begin{enumerate}
\item Bagging\\
Bagging applies the same individual classification algorithm (i.e., the base classifier) to different bootstrap samples of the training set~\citep{breiman1996bagging}. It aims at improving unstable estimations by reducing variance, while slightly increasing bias,  for a given base classifier. The outputs of single classifiers are aggregated to calculate the final output using a majority voting rule.  Given $D$ as a training set, bagging produces $m$ bootstrap samples with replacement $D_{1}, D_{2},..., D_{m}$, randomly chosen from $D$, of size $n$. For each bootstrap sample $D_{i}$, a single classification algorithm $cls_{i}$ is trained by utilizing the same classifier. To predict a test sample $w$, bagging feeds the samples to the single classifiers to obtain $m$ predictions $O=\left\{ cls_{1}(w), \ldots, cls_{m}(w) \right\}$ and chooses the predicted label as the most frequent in $O$. We use the bagging algorithm implemented in Weka. The number of bags $m$ is set to 10, while the size of each random sample $n$ is 100\%.        

\item Boosting\\
Boosting trains a set of classifiers sequentially and aggregates their results for final prediction by imposing that later classifiers pay more attention to the classification errors made by earlier learners. Many implementations of boosting exist. In this work we use the one considered by many as the most popular boosting algorithm, i.e., AdaBoost~\citep{freund1997decision}. Let $D$ be the input of boosting,  $cls$ a base classifier, and $R$ the number of learning rounds. The procedure of boosting can be described as follows: (i) applying individual classifier $cls$ to the original dataset $D$, (ii) determining the weight of the samples in $D$ such that they are inversely proportional to the classification error of $cls^{r}$, where $r$ is the current round, (iii) if $r \neq R$, then increase $r$ of one unit and go back to step (i). Given a test sample $w$ and the predictions $O=\{cls^1(w), \ldots, cls^R(w) \}$, the predicted label is chosen as the most frequent in $O$. The AdaBoostM1 algorithm available in Weka is considered in the experiment with the number of weak models to create $M$  set to 10.

\item Random Subspace\\
Random subspace uses several feature subsets to train models using the base classifier~\citep{ho1998random}. It is, therefore, a parallel algorithm in which individual classifiers are trained independently using different features. Let $F$ be the number of features generated from a dataset, $d$ be an integer, with $d < F$, and $M$ be the ensemble size, i.e., the number of base classifiers that will be trained.  Random subspace trains $M$ models $cls_1, \ldots, cls_M$ using some base learning algorithms, where for each model only $d$ features are randomly selected. Given a test sample $w$ and the predictions $O=\left\{ cls_1(w), \ldots, cls_{M}(w)\right\}$, the predicted labels for $w$ is chosen as the most frequent in $O$. We use the random subspace algorithm implemented in Weka with $F$ set to 50\% and the number of base learners $M$ set to 10.

\item Nested Dichotomies\\
A nested dichotomy (ND) decomposes a multi-class classification problem into a set of binary problems~\citep{frank2004ensembles}. The performance of a nested dichotomy relies on the selection of the decomposition and the choice of individual learners, i.e., base classifiers. Given a set $\mathcal{Y}=\{a_1,\ldots, a_i,\ldots,a_I \} \subseteq \mathcal{A}$ of $I$ distinct event types in an event log, a nested dichotomy is a recursive splitting $(\mathcal{Y}_{a},\mathcal{Y}_{b})$ of $\mathcal{Y}$ into pairs of disjoint and non-empty subsets~\citep{melnikov2018effectiveness}. 
More specifically, ND builds a binary tree with $I$ leaf nodes, which are uniquely labelled by the event types. For instance, Figure~\ref{fig1} shows the recursive separation of five distinct event types using four base classifiers. The first classifier $(cls_{1})$ splits event type $\#_a(e) = \gamma$ from the union of other event types $\#_a(e) \in \{ \alpha, \beta, \delta,\epsilon \}$; in a similar fashion, the second classifier $(cls_{2})$ splits event types $\#_a(e) \in  \{ \alpha, \delta \}$ from $\#_a(e) \in \{ \beta,\epsilon \}$, and so forth.

Once the hierarchy of base classifiers has been trained, a test sample $w$ can be predicted in a probabilistic manner. For each leaf node, a classification probability $p_i(w)$ of $\#_a(e) = a_i$ is calculated by multiplying the probabilities along the path from the root of the tree to the leaf with $\#_a(e) = a_i$. The event type $a_i$ with highest probability $p_i(w)$ is chosen as the predicted one for a test sample $w(\cdot)$. In this paper, we employ the implementation of ND algorithm supplied by Weka, where the number of base classifiers is set to 10.

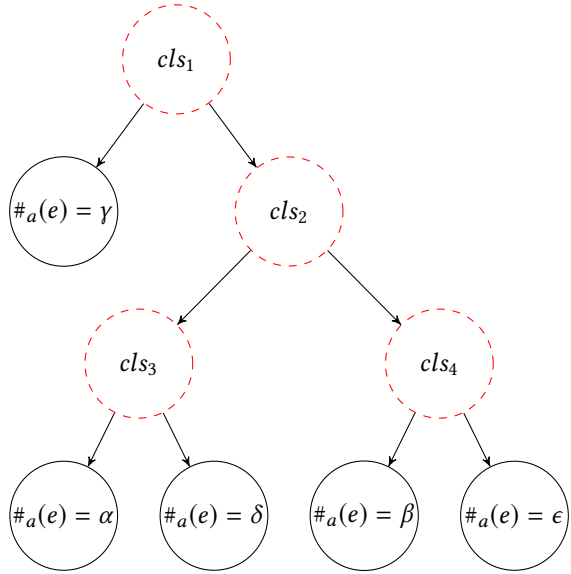
\begin{figure}[ht!]
\centering
\begin{tikzpicture}[->,>=stealth']
\node[arn_r]{$cls_{1}$}
  child{node[arn_n]{$\#_a(e) = \gamma$}}
  child{node[arn_r]{$cls_{2}$}
     child{node[arn_r]{$cls_{3}$}
        child{node[arn_n]{$\#_a(e) = \alpha$}}
        child{node[arn_n]{$\#_a(e) = \delta$}}}
     child{node[arn_r]{$cls_{4}$}
        child{node[arn_n]{$\#_a(e) = \beta$}}
        child{node[arn_n]{$\#_a(e) = \epsilon$}}}};
\end{tikzpicture}
\caption{\label{fig1} Illustration of a recursive splitting in a nested dichotomy~\citep{melnikov2018effectiveness}} 
\end{figure}

\item Dagging\\
Dagging creates several disjoint samples (instead of bootstrap in bagging) and feeds each sample of data into a copy of the given of single classifier~\citep{ting1997stacking}. 
Given a dataset $D$, it randomly samples $D$ into $M$ disjoint partitions without replacement of size $I$. Then, it trains $M$ models $cls_{m}$ using some base classifier. To classify a test instance $w$, the predictions are made via averaging in $O=\left\{ cls_1(w), \ldots, cls_{M}(w)\right\}$. In the experiment, we consider the implementation of dagging provided by Weka with number M set to 10.
\end{enumerate}

\subsection{Validation and performance significance tests}
Repeated hold-out is chosen as a validation procedure for the experiments. At each iteration, the training samples are drawn from dataset $D$ without replacement in a specified percentage (67\%), while the remaining samples are used for testing. The procedure is then repeated 30 times to reduce variations in the random splits.

We use two different statistical tests to assess whether the performance differences among the considered classifiers are significant. First, we compare all the classification algorithms using the Friedman test. The Friedman test~\citep{friedman1937use} assigns a different rank to each classifier for a given dataset in a performance ascending way, i.e., such that the the best performer is assigned the rank 1. Next, an all inclusive Friedman $p$-value is adopted  to detect whether at least one of the classifiers performs significantly differently than the others. This test is considered to reject the null hypothesis that performance differences among all classifiers for a given dataset are not significant. 

When the Friedman test rejects the null hypothesis, two pair-wise tests, i.e., Friedman posthoc~\citep{friedman1940comparison} and Rom~\citep{rom1990sequentially} tests with the corresponding $p$-value adjustment are applied for multiple comparisons among the classifier performances. Two alternatives for pairwise comparisons are available, i.e., comparison with a control and all pairwise comparisons. In this paper, we consider the top ranked classifier as the control classifier. Friedman and Rom posthoc are chosen since they are not complex while still powerful procedures when the alternatives to compare are greater than 5, such as in our case.

\section{Experimental Results and Discussion}
\label{expres}

This section presents the results obtained from the experiments. The complete experimental results and the datasets are publicly available to foster reproducibillity in future research\footnote{http://bit.ly/ensembleprocessmonitoring}. 

The accuracy achieved by each classifier individually and using different ensemble schemes is shown in Table~\ref{tab5} - \ref{tab10}, which are reported in the Appendix~\ref{perf} for maintaining the readability of the manuscript. The classifiers ranked first, second, and third in these tables are highlighted using bold, underlined, and italic fonts, respectively.
Tree-based classifiers emerge as the best base classifiers for both groups of event logs. In particular, 
DT achieves a higher average rank in both groups. This implies that DT is a stable classifier, that is, there is no substantial difference in the performance obtained using DT as an individual classifier or as the base classifier in an ensemble. The other top-4 performers are C-DT, DTNB, JRip, and NB. The worst performers are DS, SVM, and CR. 
The poor performance of SVM may be unexpected since SVM-based classifiers show remarkable performance in several application domains. However, this result is in line with the experiments in~\cite{abellan2017comparative,teinemaa2019outcome,tama2019empirical} using credit scoring datasets, where SVM was also the worst classifier. The poor performance of SVM classifiers, in this case, may be due to a poor choice and parameterization of the kernel functions. A different parameterization may be in fact needed for each different event log.


More in detail, the results of Table~\ref{tab5} - \ref{tab10} can be summarised as follows:

\begin{itemize}
\item DT wins the benchmark in almost all ensemble schemes and event logs groups, except when it is placed in boosting. 
\item Generally speaking, C-DT is the second best performing classifier over any ensemble schemes and any variability level of event log datasets. 
\item Compared to JRip, DTNB is superior in any ensemble schemes, except dagging with high variability event logs. For overall performance, DTNB occupies the third best performing classifier in our experiment. 
\item The best result for JRip is obtained when it is placed in nested dichotomies or boosting.
\end{itemize}

Note that the overall results of Table~\ref{tab5} - \ref{tab10} confirm the ones of our previously published benchmark regarding the level of accuracy achieved. The top-performing classifiers for low variability event logs achieve an average accuracy of around 85\%, while this average drops to around 65\% for high variability logs. This confirms that high variability logs are significantly more challenging for this prediction task than low variability ones and also shows that ensemble schemes do not help to improve the performance of classifiers on high variability logs.

In order to answer the research question RQ1, regarding the impact of the size of the window $l$ for encoding on the classifier performance, Table~\ref{tab11} shows, for the top-2 ranked classifiers in each setting (individual and ensemble schemes), the relative percentage performance difference on different event logs, using $l=3$ as a baseline. The results should be interpreted by considering, as an additional characterisation of the data at hand, the number of samples obtained from the application of the window-based encoding for different values of $l$, which is shown in Table~\ref{tab:window size}. For some event logs, in fact, the number of samples available decreases dramatically with the increase of $l$, which makes the models obtained unreliable and prone to overfitting. For instance, for the Road Traffic event log, the number of samples available decreases by 92\% (from 10,042 to 767 samples) when the window size increases from $l=3$ to $l=5$.

Table~\ref{tab:qualitative} summarises, from a qualitative standpoint, the results obtained for different window sizes in respect of three characteristics of an event log: (i) the number of samples for larger values of $l$, (ii) the event log variability and (iii) the mean/median trace length. Based on Table~\ref{tab:qualitative}, we highlight the following insights:

\begin{itemize}
    \item The choice of the optimal window size does not depend on the variability of an event log, but the optimal window size should be assessed individually for each event log;
    \item Increasing the window size may reduce drastically the number of samples available to train a model, so the performance obtained should always be assessed in term of its reliability based on the number of samples available;
    \item Despite the caveats highlighted above, a lower window size ($l=3,4$) appears to be a safer choice that is likely to perform satisfactorily in most cases. However, in case of large logs with longer traces, e.g., Sepsis and BPIC 2013, we suggest to test a larger set of window sizes to find an optimal one. For instance, in the case of Sepsis, $l=4$ or $l=5$ lead to the best performance in all ensemble schemes, whereas in the case of BPIC 2013, the baseline $l=3$ is the optimal choice in all schemes except Random Subspace. 
\end{itemize}

\begin{table}
\caption{Relative differences (\%) of the best two classifiers w.r.t different window sizes, where size $l=3$ is the baseline. For example, C-DT performance on the Helpdesk event log is 13.85\% higher with $l=4$ than with $l=3$.}
\label{tab11}
\centering
\resizebox{1\textwidth}{!}{
\begin{tabular}{l|l|l|l|l|l|l|l|l|l}
\hline
Scheme                              & Classifier            & \begin{tabular}[c]{@{}l@{}}Window\\ baseline $l$\end{tabular}      & \begin{tabular}[c]{@{}l@{}}Window\\ size $l$\end{tabular} & Helpdesk &  \begin{tabular}[c]{@{}l@{}}Hospital\\ Billing\end{tabular} & \begin{tabular}[c]{@{}l@{}}Road\\ Traffic\end{tabular} & BPIC2012 & \begin{tabular}[c]{@{}l@{}}BPIC2013\\   Incident\end{tabular} & Sepsis \\ \hline\hline

\multirow{10}{*}{Individual}& \multirow{5}{*}{C-DT} & \multirow{5}{*}{3} & 4&13.85&	-1.49&	-7.64&	0.95&	-4.58&	10.06 \\  
&&&5&16.05&	-4.59&	-6.89&	-5.92&	-1.67&	10.15 \\
&&&6&14.94&	-7.36&	-12.04&	-8.33&	-3.78&	8.60 \\ 
&&&7&12.32&	-8.13&	-14.51&	-15.70&	-2.87&	7.75\\
&&&8&-0.71&	-5.21&	-24.90&	-10.85&	-3.98&	2.48\\
\cline{2-10}

& \multirow{5}{*}{DT}& \multirow{5}{*}{3} & 4&11.40&	-1.51&	-7.83&	0.91&	-4.51&	10.30 \\
&&&5&12.71&	-4.66&	-6.27&	-6.08&	-1.75&	11.29 \\ 
&&&6&12.53&	-7.28&	-11.37&	-7.06&	-4.21&	9.66 \\ 
&&&7&17.27&	-7.79&	-16.17&	-11.14&	-2.98&	8.35\\
&&&8&15.10&	-4.69&	-27.77&	-8.09&	-4.09&	2.63\\\hline

\multirow{10}{*}{Bagging}& \multirow{5}{*}{C-DT} & \multirow{5}{*}{3} & 4 &11.98&	-1.45&	-7.80&	0.87&	-4.63&	11.61 \\ 
&&&5&14.63&	-4.64&	-7.35&	-6.17&	-1.32&	12.53 \\ 
&&&6&15.41&	-7.38&	-10.52&	-8.10&	-3.75&	10.35 \\ 
&&&7&15.56&	-7.77&	-13.49&	-13.85&	-2.46&	9.17\\
&&&8&15.29&	-5.11&	-19.87&	-13.75&	-3.39&	4.95\\
\cline{2-10} 

& \multirow{5}{*}{DT}   & \multirow{5}{*}{3} & 4&13.37&	-1.37&	-8.01&	0.74&	-5.68&	11.76 \\
&&&5&13.75&	-4.70&	-6.52&	-6.30&	-2.11&	14.38 \\ 
&&&6&18.66&	-7.09&	-9.01&	-8.18&	-4.04&	12.08 \\ 
&&&7&20.46&	-7.56&	-12.33&	-13.62&	-3.04&	12.08\\
&&&8&21.86&	-4.40&	-27.86&	-14.40&	-4.39&	7.32\\\hline

\multirow{10}{*}{Boosting}& \multirow{5}{*}{C-DT} & \multirow{5}{*}{3} & 4 &13.69&	-1.51&	-7.67&	1.03&	-4.34&	9.62 \\
&&&5& 16.09&	-4.65&	-5.99&	-5.74&	-1.39&	10.25\\ 
&&&6&14.92&	-7.22&	-11.12&	-8.70&	-3.55&	8.60 \\ 
&&&7&14.15&	-7.91&	-13.48&	-17.21&	-2.73&	7.17\\
&&&8&8.46&	-5.08&	-17.33&	-14.50&	-3.80&	3.41\\
\cline{2-10} 

& \multirow{5}{*}{DT}   & \multirow{5}{*}{3} & 4& 9.07&	-1.38&	-8.09&	0.71&	-5.58&	8.93\\
&&&5&10.14&	-1.94&	-7.92&	-5.94&	-3.91&	8.15 \\ 
&&&6&8.32&	-5.37&	-9.92&	-7.03&	-6.73&	4.63 \\ 
&&&7&10.05&	-6.52&	-13.47&	-10.95&	-5.90&	4.41\\
&&&8&11.49&	-3.77&	-16.40&	-5.52&	-8.24&	-2.35\\
\hline

\multirow{10}{*}{Random Subspace}    & \multirow{5}{*}{C-DT} & \multirow{5}{*}{3} & 4 &14.42&	-0.47&	-6.94&	1.30&	-0.33&	12.97 \\ 
&&&5&17.21&	-4.35&	-10.69&	-5.49&	1.88&	12.87 \\ 
&&&6&19.11&	-7.13&	-9.51&	-7.85&	0.98&	12.55 \\ 
&&&7&14.38&	-6.98&	-12.24&	-16.41&	0.66&	8.11\\
&&&8&18.17&	-3.52&	-18.73&	-10.67&	0.38&	7.80\\
\cline{2-10} 
                
& \multirow{5}{*}{DT}   & \multirow{5}{*}{3} & 4& 12.86&	-0.37&	-7.03&	1.30&	0.72&	14.35\\
&&&5&15.18&	-4.44&	-8.50&	-5.78&	3.24&	14.88 \\ 
&&&6&18.08&	-7.22&	-8.84&	-7.54&	1.76&	16.23 \\ 
&&&7&18.34&	-7.32&	-14.33&	-12.40&	1.54&	10.07\\
&&&8&17.92&	-3.10&	-23.23&	-10.08&	1.29&	11.46\\
\hline

\multirow{10}{*}{Nested Dichotomies} & \multirow{5}{*}{C-DT} & \multirow{4}{*}{3} & 4 &10.51&	-1.53&	-7.84&	0.89&	-4.94&	9.78 \\ 
&&&5&13.39&	-4.68&	-7.82&	-7.41&	-1.96&	11.02 \\ 
&&&6&16.46&	-7.41&	-18.07&	-10.44&	-3.71&	7.99 \\ 
&&&7&9.84&	-8.20&	-21.13&	-13.06&	-2.66&	6.52\\
&&&8&4.39&	-4.82&	-27.40&	-14.19&	-3.53&	2.70\\
\cline{2-10} 

& \multirow{5}{*}{DT}   & \multirow{5}{*}{3} & 4&12.37&	-1.59&	-7.93&	0.70&	-4.29&	10.25 \\ 
&&&5&13.86&	-4.70&	-7.24&	-6.54&	-3.30&	11.22 \\ 
&&&6&14.46&	-7.60&	-16.37&	-7.45&	-3.74&	9.00 \\ 
&&&7&18.90&	-8.07&	-17.37&	-11.54&	-3.54&	6.90\\
&&&8&17.96&	-5.19&	-26.40&	-10.39&	-5.81&	1.90\\
\hline
\multicolumn{10}{c}{\textit{Continued on next page}}\\
\hline
\end{tabular}
}
\end{table}

\begin{table}
\centering
\resizebox{1\textwidth}{!}{
\begin{tabular}{l|l|l|l|l|l|l|l|l|l}
\hline
\multicolumn{10}{c}{\textit{Continued on previous page}}\\
\hline
Scheme                              & Classifier            & \begin{tabular}[c]{@{}l@{}}Window\\ baseline $l$\end{tabular}      & \begin{tabular}[c]{@{}l@{}}Window\\ size $l$\end{tabular} & Helpdesk &  \begin{tabular}[c]{@{}l@{}}Hospital\\ Billing\end{tabular} & \begin{tabular}[c]{@{}l@{}}Road\\ Traffic\end{tabular} & BPIC2012 & \begin{tabular}[c]{@{}l@{}}BPIC2013\\   Incident\end{tabular} & Sepsis \\ 
\hline\hline

\multirow{10}{*}{Dagging} & \multirow{5}{*}{C-DT} & \multirow{5}{*}{3} & 4 &12.86&	-1.83&	-5.93&	1.32&	-4.61&	3.02 \\
&&&5&12.17&	-5.09&	-11.02&	-5.79&	-1.63&	3.77 \\ 
&&&6&9.61&	-8.28&	-26.53&	-15.42&	-3.39&	2.90 \\ 
&&&7&-13.32&	-8.83&	-47.67&	-15.86&	-2.39&	4.72\\
&&&8&-20.48&	-6.28&	-46.93&	-10.56&	-3.20&	3.19\\
\cline{2-10} 

& \multirow{5}{*}{DT}   & \multirow{5}{*}{3} & 4&11.41&	-1.50&	-7.92&	1.35&	-4.53&	6.94 \\
&&&5&13.91&	-4.71&	-11.49&	-5.63&	-1.50&	7.63 \\ 
&&&6&17.01&	-7.55&	-20.41&	-7.78&	-3.71&	4.79 \\ 
&&&7&2.61&	-8.22&	-20.63&	-13.89&	-2.97&	4.22\\
&&&8&-26.14&	-5.33&	-31.40&	-10.52&	-3.49&	1.98\\
\hline
\end{tabular}
}
\end{table}

\begin{table}[]
\centering
\caption{\label{tab:window size} Sample size after encoding for different window size $l$}
\begin{tabular}{l|l|l|l|l|l|l}
\hline
\multirow{2}{*}{Event log} & \multicolumn{6}{c}{Sample size}               \\
\cline{2-7}
                           & $l=3$ & $l=4$ & $l=5$ & $l=6$ & $l=7$ & $l=8$ \\
\hline
\hline
Helpdesk                   & 2,477  & 1,117  & 490   & 227   & 101   & 56    \\
Hospital   Billing         & 24,337 & 17,066 & 9,877  & 4,936  & 3,696  & 2,887  \\
Road Traffic               & 10,042 & 5,402  & 767   & 163   & 101   & 49    \\
\hline
Sepsis                     & 12,043 & 11,029 & 10,016 & 9,064  & 8,144  & 7,251  \\
BPIC2013                   & 42,883 & 37,101 & 32,193 & 27,709 & 23,964 & 20,789 \\
BPIC2012                   & 6,579  & 3,657  & 1,911  & 952   & 444   & 197 \\ 
\hline
\end{tabular}
\end{table}

\newcolumntype{L}[1]{>{\raggedright\let\newline\\\arraybackslash\hspace{0pt}}m{#1}}
\newcolumntype{C}[1]{>{\centering\let\newline\\\arraybackslash\hspace{0pt}}m{#1}}
\newcolumntype{R}[1]{>{\raggedleft\let\newline\\\arraybackslash\hspace{0pt}}m{#1}}
\begin{table}[]
\caption{\label{tab:qualitative}Qualitative evaluation of optimal encoding window length $l$ for different event logs.}
\label{tab11.5}
\centering
\begin{tabular}{ L{2.4cm} | C{1.8cm} | C{1.8cm} | C{1.8cm} | L{4cm} }
\hline
Event log                         & Number of samples for larger $l$                    & Event Log Variability          & Median/Mean Trace Length         & Optimal value(s) of $l$ \\
\hline
\hline
\multirow{2}{*}{Helpdesk}         & \multirow{2}{*}{Low} & \multirow{2}{*}{Low} & \multirow{2}{*}{Low} & No clear optimal value;      \\
                                  &                      &                      &                      & unreliable results for larger values of $l$.   \\
                                  \hline
\multirow{2}{*}{Hospital Billing} & \multirow{2}{*}{High} & \multirow{2}{*}{Low} & \multirow{2}{*}{Low} & $l=3$      \\
                                  &                      &                      &                      & Reliable results for all values of $l$.   \\
                                  \hline
\multirow{2}{*}{Road Traffic}     & \multirow{2}{*}{Low} & \multirow{2}{*}{Low} & \multirow{2}{*}{Low} & $l=3$      \\
                                  &                      &                      &                      & Unreliable results for larger values of $l$.   \\
                                  \hline
\multirow{2}{*}{Sepsis}           & \multirow{2}{*}{High} & \multirow{2}{*}{High} & \multirow{2}{*}{High} & $l=4,5$      \\
                                  &                      &                      &                      & Reliable results for all values of $l$.   \\
                                  \hline
\multirow{2}{*}{BPIC2013}         & \multirow{2}{*}{High} & \multirow{2}{*}{High} & \multirow{2}{*}{High} & $l=3$      \\
                                  &                      &                      &                      & Reliable results for all values of $l$.   \\
                                  \hline
\multirow{2}{*}{BPIC2012}         & \multirow{2}{*}{Low} & \multirow{2}{*}{High} & \multirow{2}{*}{Low} & $l=4$      \\
                                  &                      &                      &                      & Unreliable results for larger values of $l$.  \\
                                  \hline
\end{tabular}
\end{table}

\begin{figure}
\captionsetup[subfigure]{font=footnotesize}
\subcaptionbox{}[1\textwidth]{
\centering
\resizebox{.7\textwidth}{!}{
\begin{tikzpicture}
\begin{axis}[
legend pos=south east,
ylabel={Average accuracy (\%)},
ymin=15,ymax=95,
symbolic x coords={SVM,DS,CR,RT,k-NN,NB,OneR,BN,JRip,C-DT,DTNB,DT},
xtick=data,
ymajorgrids=true,
xmajorgrids=true,
grid style=dashed,
]
\addplot[color=black,mark=square,mark size=2.5pt] table[x=class,y=Individual,col sep=comma]{low3.csv};
\addplot[color=red,mark=asterisk,mark size=2.5pt] table[x=class,y=Bagging,col sep=comma]{low3.csv};
\addplot[color=green,mark=triangle,mark size=2.5pt] table[x=class,y=Boosting,col sep=comma]{low3.csv};
\addplot[color=blue,mark=diamond,mark size=2.5pt] table[x=class,y=Random_Subspace,col sep=comma]{low3.csv};
\addplot[color=cyan,mark=o,mark size=2.5pt] table[x=class,y=Nested_Dichotomies,col sep=comma]{low3.csv};
\addplot[color=magenta,mark=x,mark size=2.5pt] table[x=class,y=Dagging,col sep=comma]{low3.csv};
\legend{Individual,Bagging,Boosting,Random Subspace,Nested Dichotomies,Dagging}
\end{axis}
\end{tikzpicture}}
}

\captionsetup[subfigure]{font=footnotesize}
\subcaptionbox{}[1\textwidth]{
\centering
\resizebox{.7\textwidth}{!}{
\begin{tikzpicture}
\begin{axis}[
legend pos=south east,
ylabel={Average accuracy (\%)},
ymin=15,ymax=95,
symbolic x coords={SVM,DS,CR,k-NN,RT,NB,JRip,BN,OneR,DTNB,DT,C-DT
},
xtick=data,
ymajorgrids=true,
xmajorgrids=true,
grid style=dashed,
]
\addplot[color=black,mark=square,mark size=2.5pt] table[x=class,y=Individual,col sep=comma]{high3.csv};
\addplot[color=red,mark=asterisk,mark size=2.5pt] table[x=class,y=Bagging,col sep=comma]{high3.csv};
\addplot[color=green,mark=triangle,mark size=2.5pt] table[x=class,y=Boosting,col sep=comma]{high3.csv};
\addplot[color=blue,mark=diamond,mark size=2.5pt] table[x=class,y=Random_Subspace,col sep=comma]{high3.csv};
\addplot[color=cyan,mark=o,mark size=2.5pt] table[x=class,y=Nested_Dichotomies,col sep=comma]{high3.csv};
\addplot[color=magenta,mark=x,mark size=2.5pt] table[x=class,y=Dagging,col sep=comma]{high3.csv};
\legend{Individual,Bagging,Boosting,Random Subspace,Nested Dichotomies,Dagging}
\end{axis}
\end{tikzpicture}}
}
\caption{\label{window3}Average accuracy of all classifiers over low (a) and high (b) variability for $l=3$}
\end{figure}

\begin{figure}
\captionsetup[subfigure]{font=footnotesize}
\subcaptionbox{}[1\textwidth]{
\centering
\resizebox{.7\textwidth}{!}{
\begin{tikzpicture}
\begin{axis}[
legend pos=south east,
ylabel={Average accuracy (\%)},
ymin=15,ymax=95,
symbolic x coords={SVM,CR,DS,RT,k-NN,NB,BN,OneR,DTNB,JRip,C-DT,DT
},
xtick=data,
ymajorgrids=true,
xmajorgrids=true,
grid style=dashed,
]
\addplot[color=black,mark=square,mark size=2.5pt] table[x=class,y=Individual,col sep=comma]{low4.csv};
\addplot[color=red,mark=asterisk,mark size=2.5pt] table[x=class,y=Bagging,col sep=comma]{low4.csv};
\addplot[color=green,mark=triangle,mark size=2.5pt] table[x=class,y=Boosting,col sep=comma]{low4.csv};
\addplot[color=blue,mark=diamond,mark size=2.5pt] table[x=class,y=Random_Subspace,col sep=comma]{low4.csv};
\addplot[color=cyan,mark=o,mark size=2.5pt] table[x=class,y=Nested_Dichotomies,col sep=comma]{low4.csv};
\addplot[color=magenta,mark=x,mark size=2.5pt] table[x=class,y=Dagging,col sep=comma]{low4.csv};
\legend{Individual,Bagging,Boosting,Random Subspace,Nested Dichotomies,Dagging}
\end{axis}
\end{tikzpicture}}
}

\captionsetup[subfigure]{font=footnotesize}
\subcaptionbox{}[1\textwidth]{
\centering
\resizebox{.7\textwidth}{!}{
\begin{tikzpicture}
\begin{axis}[
legend pos=south east,
ylabel={Average accuracy (\%)},
ymin=15,ymax=95,
symbolic x coords={SVM,CR,DS,k-NN,RT,JRip,OneR,NB,BN,DTNB,DT,C-DT
},
xtick=data,
ymajorgrids=true,
xmajorgrids=true,
grid style=dashed,
]
\addplot[color=black,mark=square,mark size=2.5pt] table[x=class,y=Individual,col sep=comma]{high4.csv};
\addplot[color=red,mark=asterisk,mark size=2.5pt] table[x=class,y=Bagging,col sep=comma]{high4.csv};
\addplot[color=green,mark=triangle,mark size=2.5pt] table[x=class,y=Boosting,col sep=comma]{high4.csv};
\addplot[color=blue,mark=diamond,mark size=2.5pt] table[x=class,y=Random_Subspace,col sep=comma]{high4.csv};
\addplot[color=cyan,mark=o,mark size=2.5pt] table[x=class,y=Nested_Dichotomies,col sep=comma]{high4.csv};
\addplot[color=magenta,mark=x,mark size=2.5pt] table[x=class,y=Dagging,col sep=comma]{high4.csv};
\legend{Individual,Bagging,Boosting,Random Subspace,Nested Dichotomies,Dagging}
\end{axis}
\end{tikzpicture}}
}
\caption{\label{window4}Average accuracy of all classifiers over low (a) and high (b) variability for $l=4$}
\end{figure}

\begin{figure}
    \centering
    \captionsetup[subfigure]{font=footnotesize}
\subcaptionbox{}[1\textwidth]{
\centering
\resizebox{.7\textwidth}{!}{
\begin{tikzpicture}
\begin{axis}[
legend pos=south east,
ylabel={Average accuracy (\%)},
ymin=15,ymax=95,
symbolic x coords={SVM,DS,CR,NB,BN,OneR,RT,k-NN,DTNB,JRip,C-DT,DT},
xtick=data,
ymajorgrids=true,
xmajorgrids=true,
grid style=dashed,
]
\addplot[color=black,mark=square,mark size=2.5pt] table[x=class,y=Individual,col sep=comma]{low5.csv};
\addplot[color=red,mark=asterisk,mark size=2.5pt] table[x=class,y=Bagging,col sep=comma]{low5.csv};
\addplot[color=green,mark=triangle,mark size=2.5pt] table[x=class,y=Boosting,col sep=comma]{low5.csv};
\addplot[color=blue,mark=diamond,mark size=2.5pt] table[x=class,y=Random_Subspace,col sep=comma]{low5.csv};
\addplot[color=cyan,mark=o,mark size=2.5pt] table[x=class,y=Nested_Dichotomies,col sep=comma]{low5.csv};
\addplot[color=magenta,mark=x,mark size=2.5pt] table[x=class,y=Dagging,col sep=comma]{low5.csv};
\legend{Individual,Bagging,Boosting,Random Subspace,Nested Dichotomies,Dagging}
\end{axis}
\end{tikzpicture}}
}
\captionsetup[subfigure]{font=footnotesize}
\subcaptionbox{}[1\textwidth]{
\centering
\resizebox{.7\textwidth}{!}{
\begin{tikzpicture}
\begin{axis}[
legend pos=south east,
ylabel={Average accuracy (\%)},
ymin=15,ymax=95,
symbolic x coords={SVM,DS,CR,k-NN,RT,JRip,OneR,NB,BN,DTNB,DT,C-DT
},
xtick=data,
ymajorgrids=true,
xmajorgrids=true,
grid style=dashed,
]
\addplot[color=black,mark=square,mark size=2.5pt] table[x=class,y=Individual,col sep=comma]{high5.csv};
\addplot[color=red,mark=asterisk,mark size=2.5pt] table[x=class,y=Bagging,col sep=comma]{high5.csv};
\addplot[color=green,mark=triangle,mark size=2.5pt] table[x=class,y=Boosting,col sep=comma]{high5.csv};
\addplot[color=blue,mark=diamond,mark size=2.5pt] table[x=class,y=Random_Subspace,col sep=comma]{high5.csv};
\addplot[color=cyan,mark=o,mark size=2.5pt] table[x=class,y=Nested_Dichotomies,col sep=comma]{high5.csv};
\addplot[color=magenta,mark=x,mark size=2.5pt] table[x=class,y=Dagging,col sep=comma]{high5.csv};
\legend{Individual,Bagging,Boosting,Random Subspace,Nested Dichotomies,Dagging}
\end{axis}
\end{tikzpicture}}
}
\caption{\label{window5}Average accuracy of all classifiers over low (a) and high (b) variability for $l=5$}
\end{figure}

\begin{figure}
    \centering
    \captionsetup[subfigure]{font=footnotesize}
\subcaptionbox{}[1\textwidth]{
\centering
\resizebox{.7\textwidth}{!}{
\begin{tikzpicture}
\begin{axis}[
legend pos=south east,
ylabel={Average accuracy (\%)},
ymin=15,ymax=95,
symbolic x coords={SVM,DS,CR,BN,NB,OneR,RT,JRip,k-NN,DTNB,C-DT,DT},
xtick=data,
ymajorgrids=true,
xmajorgrids=true,
grid style=dashed,
]
\addplot[color=black,mark=square,mark size=2.5pt] table[x=class,y=Individual,col sep=comma]{low6.csv};
\addplot[color=red,mark=asterisk,mark size=2.5pt] table[x=class,y=Bagging,col sep=comma]{low6.csv};
\addplot[color=green,mark=triangle,mark size=2.5pt] table[x=class,y=Boosting,col sep=comma]{low6.csv};
\addplot[color=blue,mark=diamond,mark size=2.5pt] table[x=class,y=Random_Subspace,col sep=comma]{low6.csv};
\addplot[color=cyan,mark=o,mark size=2.5pt] table[x=class,y=Nested_Dichotomies,col sep=comma]{low6.csv};
\addplot[color=magenta,mark=x,mark size=2.5pt] table[x=class,y=Dagging,col sep=comma]{low6.csv};
\legend{Individual,Bagging,Boosting,Random Subspace,Nested Dichotomies,Dagging}
\end{axis}
\end{tikzpicture}}
}

\captionsetup[subfigure]{font=footnotesize}
\subcaptionbox{}[1\textwidth]{
\centering
\resizebox{.7\textwidth}{!}{
\begin{tikzpicture}
\begin{axis}[
legend pos=south east,
ylabel={Average accuracy (\%)},
ymin=15,ymax=95,
symbolic x coords={SVM,CR,DS,k-NN,RT,JRip,OneR,NB,BN,DTNB,C-DT,DT
},
xtick=data,
ymajorgrids=true,
xmajorgrids=true,
grid style=dashed,
]
\addplot[color=black,mark=square,mark size=2.5pt] table[x=class,y=Individual,col sep=comma]{high6.csv};
\addplot[color=red,mark=asterisk,mark size=2.5pt] table[x=class,y=Bagging,col sep=comma]{high6.csv};
\addplot[color=green,mark=triangle,mark size=2.5pt] table[x=class,y=Boosting,col sep=comma]{high6.csv};
\addplot[color=blue,mark=diamond,mark size=2.5pt] table[x=class,y=Random_Subspace,col sep=comma]{high6.csv};
\addplot[color=cyan,mark=o,mark size=2.5pt] table[x=class,y=Nested_Dichotomies,col sep=comma]{high6.csv};
\addplot[color=magenta,mark=x,mark size=2.5pt] table[x=class,y=Dagging,col sep=comma]{high6.csv};
\legend{Individual,Bagging,Boosting,Random Subspace,Nested Dichotomies,Dagging}
\end{axis}
\end{tikzpicture}}
}
\caption{\label{window5}Average accuracy of all classifiers over low (a) and high (b) variability for $l=6$}
\end{figure}

\begin{figure}
    \centering
    \captionsetup[subfigure]{font=footnotesize}
\subcaptionbox{}[1\textwidth]{
\centering
\resizebox{.7\textwidth}{!}{
\begin{tikzpicture}
\begin{axis}[
legend pos=south east,
ylabel={Average accuracy (\%)},
ymin=15,ymax=95,
symbolic x coords={SVM,DS,CR,OneR,BN,NB,k-NN,RT,C-DT,JRip,DTNB,DT
},
xtick=data,
ymajorgrids=true,
xmajorgrids=true,
grid style=dashed,
]
\addplot[color=black,mark=square,mark size=2.5pt] table[x=class,y=Individual,col sep=comma]{low7.csv};
\addplot[color=red,mark=asterisk,mark size=2.5pt] table[x=class,y=Bagging,col sep=comma]{low7.csv};
\addplot[color=green,mark=triangle,mark size=2.5pt] table[x=class,y=Boosting,col sep=comma]{low7.csv};
\addplot[color=blue,mark=diamond,mark size=2.5pt] table[x=class,y=Random_Subspace,col sep=comma]{low7.csv};
\addplot[color=cyan,mark=o,mark size=2.5pt] table[x=class,y=Nested_Dichotomies,col sep=comma]{low7.csv};
\addplot[color=magenta,mark=x,mark size=2.5pt] table[x=class,y=Dagging,col sep=comma]{low7.csv};
\legend{Individual,Bagging,Boosting,Random Subspace,Nested Dichotomies,Dagging}
\end{axis}
\end{tikzpicture}}
}
\captionsetup[subfigure]{font=footnotesize}
\subcaptionbox{}[1\textwidth]{
\centering
\resizebox{.7\textwidth}{!}{
\begin{tikzpicture}
\begin{axis}[
legend pos=south east,
ylabel={Average accuracy (\%)},
ymin=15,ymax=95,
symbolic x coords={SVM,CR,DS,k-NN,RT,JRip,OneR,NB,BN,C-DT,DTNB,DT},
xtick=data,
ymajorgrids=true,
xmajorgrids=true,
grid style=dashed,
]
\addplot[color=black,mark=square,mark size=2.5pt] table[x=class,y=Individual,col sep=comma]{high7.csv};
\addplot[color=red,mark=asterisk,mark size=2.5pt] table[x=class,y=Bagging,col sep=comma]{high7.csv};
\addplot[color=green,mark=triangle,mark size=2.5pt] table[x=class,y=Boosting,col sep=comma]{high7.csv};
\addplot[color=blue,mark=diamond,mark size=2.5pt] table[x=class,y=Random_Subspace,col sep=comma]{high7.csv};
\addplot[color=cyan,mark=o,mark size=2.5pt] table[x=class,y=Nested_Dichotomies,col sep=comma]{high7.csv};
\addplot[color=magenta,mark=x,mark size=2.5pt] table[x=class,y=Dagging,col sep=comma]{high7.csv};
\legend{Individual,Bagging,Boosting,Random Subspace,Nested Dichotomies,Dagging}
\end{axis}
\end{tikzpicture}}
}

\caption{\label{window5}Average accuracy of all classifiers over low (a) and high (b) variability for $l=7$}
\end{figure}

\begin{figure}
    \centering
    \captionsetup[subfigure]{font=footnotesize}
\subcaptionbox{}[1\textwidth]{
\centering
\resizebox{.7\textwidth}{!}{
\begin{tikzpicture}
\begin{axis}[
legend pos=south east,
ylabel={Average accuracy (\%)},
ymin=15,ymax=95,
symbolic x coords={SVM,CR,DS,C-DT,RT,k-NN,NB,BN,DT,JRip,DTNB,OneR
},
xtick=data,
ymajorgrids=true,
xmajorgrids=true,
grid style=dashed,
]
\addplot[color=black,mark=square,mark size=2.5pt] table[x=class,y=Individual,col sep=comma]{low8.csv};
\addplot[color=red,mark=asterisk,mark size=2.5pt] table[x=class,y=Bagging,col sep=comma]{low8.csv};
\addplot[color=green,mark=triangle,mark size=2.5pt] table[x=class,y=Boosting,col sep=comma]{low8.csv};
\addplot[color=blue,mark=diamond,mark size=2.5pt] table[x=class,y=Random_Subspace,col sep=comma]{low8.csv};
\addplot[color=cyan,mark=o,mark size=2.5pt] table[x=class,y=Nested_Dichotomies,col sep=comma]{low8.csv};
\addplot[color=magenta,mark=x,mark size=2.5pt] table[x=class,y=Dagging,col sep=comma]{low8.csv};
\legend{Individual,Bagging,Boosting,Random Subspace,Nested Dichotomies,Dagging}
\end{axis}
\end{tikzpicture}}
}
\captionsetup[subfigure]{font=footnotesize}
\subcaptionbox{}[1\textwidth]{
\centering
\resizebox{.7\textwidth}{!}{
\begin{tikzpicture}
\begin{axis}[
legend pos=south east,
ylabel={Average accuracy (\%)},
ymin=15,ymax=95,
symbolic x coords={SVM,CR,DS,k-NN,RT,NB,BN,JRip,OneR,C-DT,DT,DTNB
},
xtick=data,
ymajorgrids=true,
xmajorgrids=true,
grid style=dashed,
]
\addplot[color=black,mark=square,mark size=2.5pt] table[x=class,y=Individual,col sep=comma]{high8.csv};
\addplot[color=red,mark=asterisk,mark size=2.5pt] table[x=class,y=Bagging,col sep=comma]{high8.csv};
\addplot[color=green,mark=triangle,mark size=2.5pt] table[x=class,y=Boosting,col sep=comma]{high8.csv};
\addplot[color=blue,mark=diamond,mark size=2.5pt] table[x=class,y=Random_Subspace,col sep=comma]{high8.csv};
\addplot[color=cyan,mark=o,mark size=2.5pt] table[x=class,y=Nested_Dichotomies,col sep=comma]{high8.csv};
\addplot[color=magenta,mark=x,mark size=2.5pt] table[x=class,y=Dagging,col sep=comma]{high8.csv};
\legend{Individual,Bagging,Boosting,Random Subspace,Nested Dichotomies,Dagging}
\end{axis}
\end{tikzpicture}}
}

\caption{\label{window5}Average accuracy of all classifiers over low (a) and high (b) variability for $l=8$}
\end{figure}

In order to answer RQ2, regarding to what extent classifier ensembles result in a performance increase over individual classifiers, Figure~\ref{window3}-\ref{window5} show the average accuracy of all classifiers individually and in different ensemble schemes for low variability and high variability event logs for $l=3,4,5,6,7,8$, respectively. The implementation of ensemble methods always brings significant improvements over individual classifiers, i.e. DS, RT and CR, and for SVM (except in the dagging ensemble for large window sizes). 
This implies that these classifiers are unstable in this prediction task. Conversely, the other classifiers can be considered stable in this prediction task, since their usage in ensemble schemes does not improve their performance (but, actually, in some cases performance clearly deteriorates when using these base classifiers in an ensemble scheme).
Focusing only on the top-performers (C-DT, DTNB, DT), it can be noted that these are very stable for both low and high variability event logs.

\begin{table}
\centering
\caption{\label{tab12}Results of Friedman test for all classifier schemes}
\begin{tabular}{lllll}
\hline
Variability           & Scheme                                                         & $\chi_{F}^{2}$    & $p$-value  & $H_{0}$ rejection \\
\hline\hline
\multirow{6}{*}{Low}  & Individual                                                     & 115.89 & < 2.2E-16 & Yes          \\
                      & Bagging                                                        & 127.76 & < 2.2E-16  & Yes          \\
                      & Boosting                                                       & 128.03 & < 2.2E-16  & Yes          \\
                      & \begin{tabular}[c]{@{}l@{}}Random\\   Subspace\end{tabular}    & 115.37 & < 2.2E-16 & Yes          \\
                      & \begin{tabular}[c]{@{}l@{}}Nested\\   Dichotomies\end{tabular} & 119.66 & < 2.2E-16 & Yes          \\
                      & Dagging                                                        & 119.29 & < 2.2E-16  & Yes          \\\hline
\multirow{6}{*}{High} & Individual                                                     & 149.17 & < 2.2E-16 & Yes          \\
                      & Bagging                                                        & 134.94 & < 2.2E-16  & Yes          \\
                      & Boosting                                                       & 143.84 & < 2.2E-16  & Yes          \\
                      & \begin{tabular}[c]{@{}l@{}}Random\\   Subspace\end{tabular}    & 141.85 & < 2.2E-16 & Yes          \\
                      & \begin{tabular}[c]{@{}l@{}}Nested\\   Dichotomies\end{tabular} & 136.99 & < 2.2E-16 & Yes          \\
                      & Dagging                                                        & 122.75 & < 2.2E-16 & Yes\\\hline         
\end{tabular}
\end{table}

\begin{table}
\caption{Result of Friedman posthoc and Rom with $p$-value adjustment for low variability event logs with DT as the control algorithm. Bold indicates significance.}
\label{tab13}
\resizebox{1.06\textwidth}{!}{
\begin{tabular}{l|l|lllllllllll}
\hline
Classifier scheme& Posthoc test & C-DT     & RT     & DS     & NB     & SVM    & k-NN   & JRip   & OneR   & CR     & BN     & DTNB   \\
\hline\hline
\multirow{2}{*}{Individual} & Unadj.-$p$ &0.277& 	\textbf{0.006}& 	\textbf{0}& 	\textbf{0.012}& 	\textbf{0}& 	\textbf{0.010}& 	0.712& 	0.367& 	\textbf{0}& 	\textbf{0.007}& 	0.835\\
& $p_{Rom}$& 1& 	\textbf{0.047}& 	\textbf{0}& 	0.058& 	\textbf{0}& 	0.057& 	1& 	1& 	\textbf{0}& 	\textbf{0.047}& 	1\\\hline
                                    
\multirow{2}{*}{Bagging}& Unadj.-$p$ &0.474& 	\textbf{0.033}& 	\textbf{0}& 	\textbf{0}& 	\textbf{0}& 	\textbf{0}& 	0.380& 	\textbf{0.027}& 	\textbf{0}& 	\textbf{0}& 	0.595\\
& $p_{Rom}$ &1& 	0.132& 	\textbf{0}&	\textbf{0.003} &	\textbf{0}& 	\textbf{0.002} &	1 &	0.130&\textbf{0}& 	\textbf{0.002}&	1 \\\hline
                                    
\multirow{2}{*}{Boosting}           & Unadj.-$p$ &0.083& 	\textbf{0}&	\textbf{0}& 	\textbf{0}& 	\textbf{0}& 	\textbf{0} &	0.332 &	\textbf{0.024}& 	\textbf{0}& 	\textbf{0}& 	0.782 \\
                                    & $p_{Rom}$         & 0.246 &	\textbf{0.002}& 	\textbf{0}& 	\textbf{0}&	\textbf{0}&	\textbf{0}& 	0.663 &	0.092& 	\textbf{0} &	\textbf{0.002}& 	0.782  \\\hline
                                    
\multirow{2}{*}{Random Subspace}    & Unadj.-$p$ & 0.908 &	\textbf{0.015} &	\textbf{0}& 	\textbf{0.001} &	\textbf{0} &	\textbf{0.002}& 	0.052 &	\textbf{0.001} &	\textbf{0}& 	\textbf{0.003} &	0.212  \\
                                    & $p_{Rom}$         &0.908& 	0.060 &	\textbf{0} &	\textbf{0.006} &	\textbf{0} &	\textbf{0.009} &	0.155& 	\textbf{0.008}& 	\textbf{0}& 	\textbf{0.015}& 	0.424   \\\hline
                                    
\multirow{2}{*}{Nested Dichotomies} & Unadj.-$p$ &0.248& 	\textbf{0.025} &	\textbf{0} &\textbf{0}& \textbf{0} &	\textbf{0.017} &	0.799& 	0.257& 	\textbf{0} &	\textbf{0.001} &	1  \\
                                    & $p_{Rom}$         &0.975 &	0.122& 	\textbf{0.001} &	\textbf{0.002}& 	\textbf{0} &	0.102 &	1 &.975 &	\textbf{0} &	\textbf{0.005} &	1   \\\hline
                                    
\multirow{2}{*}{Dagging}            & Unadj.-$p$ &0.052 &	0.052& 	\textbf{0.005}& 	0.079 	&\textbf{0} &	0.579& 	0.309 &	0.392 &	\textbf{0} &	0.071 &	0.694  \\
                                    & $p_{Rom}$         &0.409& 	0.409 &	\textbf{0.045} &	0.420 &	\textbf{0} &	1 &	1 &	1& 	\textbf{0} &	0.420 &	1  \\\hline
                                    
\multicolumn{13}{l}{\footnotesize Unadj.-$p$: the $p$-value obtained by Friedman post-hoc; $p_{Rom}$: the $p$-value obtained by Rom post-hoc with adjustment.}\\
\end{tabular}
}
\end{table}

\begin{table}
\caption{Result of Friedman posthoc and Rom with $p$-value adjustment for high variability event logs with DT as the control algorithm. Bold indicates significance.}
\label{tab14}
\resizebox{1.06\textwidth}{!}{
\begin{tabular}{l|l|lllllllllll}
\hline
Classifier scheme & Posthoc test & C-DT     & RT     & DS     & NB     & SVM    & k-NN   & JRip   & OneR   & CR     & BN     & DTNB   \\\hline\hline
\multirow{2}{*}{Individual}& Unadj.-$p$ &1& 	\textbf{0}&	\textbf{0}&	\textbf{0.001} &	\textbf{0}& 	\textbf{0}& 	\textbf{0.004} &	\textbf{0.001}& \textbf{0}&	\textbf{0.006} &	0.548\\
& $p_{Rom}$        &1 &	\textbf{0}& 	\textbf{0} &	\textbf{0.004}& 	\textbf{0} &	\textbf{0}&	\textbf{0.014} &	\textbf{0.004} &	\textbf{0} &	\textbf{0.019}& 	1\\\hline
\multirow{2}{*}{Bagging}& Unadj.-$p$ &0.188 & 	\textbf{0} & 	\textbf{0} & 	\textbf{0} & 	\textbf{0} & 	\textbf{0} & 	\textbf{0.001}&  	\textbf{0} & 	\textbf{0}& 	\textbf{0.003} & 	0.746 \\
& $p_{Rom}$& 0.375 &	\textbf{0}& 	\textbf{0}&  	\textbf{0.001}&  	\textbf{0} & 	\textbf{0}&  \textbf{0.006} & 	\textbf{0}&  	\textbf{0}& 	\textbf{0.010}&  	0.746  \\\hline
\multirow{2}{*}{Boosting} & Unadj.-$p$ &0.419 & 	\textbf{0} & 	\textbf{0}& 	\textbf{0.004} & 	\textbf{0}& 	\textbf{0}& 	\textbf{0.002} & 	\textbf{0}& 	\textbf{0} & 	0.061&  	0.945 \\
& $p_{Rom}$        &0.837&  	\textbf{0} & 	\textbf{0} & 	\textbf{0.014}&  	\textbf{0}&  	\textbf{0} & 	\textbf{0.010} & 	\textbf{0.001} & 	\textbf{0}&  	0.181 & 	0.945  \\\hline
\multirow{2}{*}{Random Subspace}& Unadj.-$p$ &0.488 & 	\textbf{0} & 	\textbf{0} & 	\textbf{0.002} & 	\textbf{0}&  	\textbf{0} & 	\textbf{0.001} & 	\textbf{0} & 	\textbf{0} & 	\textbf{0.009} & 	0.764 \\
& $p_{Rom}$        & 0.976&  	\textbf{0} & 	\textbf{0}&  	\textbf{0.008}&  	\textbf{0} & 	\textbf{0} & 	\textbf{0.005} & 	\textbf{0} & 	\textbf{0}&  	\textbf{0.027} & 	0.976 \\\hline
\multirow{2}{*}{Nested Dichotomies} & Unadj.-$p$ & 0.405 & 	\textbf{0}&  	\textbf{0.002}&  	\textbf{0.005} & 	\textbf{0} & 	\textbf{0} & 	0.729 & 	\textbf{0} & 	\textbf{0} & 	\textbf{0.017} & 	0.871 \\
& $p_{Rom}$& 1 & 	\textbf{0} & 	\textbf{0.011} & 	\textbf{0.025} & 	\textbf{0}& 	\textbf{0} & 	1 & 	\textbf{0.003} & 	\textbf{0} & 	0.068 & 	1 \\\hline
\multirow{2}{*}{Dagging}& Unadj.-$p$ &0.075 & 	0.729 & 	\textbf{0} & 	\textbf{0.004} & 	\textbf{0} & 	0.204 & 	0.111&  	0.061 & 	\textbf{0} & 	\textbf{0.049} & 	0.392  \\
& $p_{Rom}$        & 0.369 & 	0.785 & 	\textbf{0} & 	\textbf{0.030} & 	\textbf{0} & 	0.603 & 	0.436 & 	0.360 & 	\textbf{0} & 	0.339 & 	0.785 \\\hline
\multicolumn{13}{l}{\footnotesize Unadj.-$p$: the $p$-value obtained by Friedman post-hoc; $p_{Rom}$: the $p$-value obtained by Rom post-hoc with adjustment.}\\
\end{tabular}
}
\end{table}

We perform significance tests for each ensemble scheme using the Friedman test. The test is carried out at the level of significance of $\alpha=0.05$. As can be seen in Table~\ref{tab12}, the difference among classifiers is highly significant ($p$-value $<$ 0.05), which means that there is at least one of the classifiers that performs significantly different than others. Therefore, the null hypothesis (all classifiers have performed equivalently) must be rejected and posthoc tests should be carried out. 

Table~\ref{tab13} and~\ref{tab14} show the pairwise comparison results of the raw (unadjusted) $p$-value using the Friedman posthoc test and all the adjusted $p$-values for each scenario that incorporates a control classifier using Rom posthoc test. We consider the best performer, DT as a control classifier and compare its performance with one of all the other classifiers considered in our benchmark. Within a tolerance of $\alpha$=0.05, we can see that when $p<$ 0.05, the classifiers are worse than the control classifier.

In summary, Table~\ref{tab13} and~\ref{tab14} show that:
\begin{itemize}
\item Regarding individual classifiers, DT (used as control) achieves statistically better performance than RT, DS, SVM, CR, and BN. 

\item When using bagging, the control classifier outperforms DS, NB, SVM, $k$-NN, CR, and BN.
\item When using boosting, some classifiers, such as RT, DS, CR, NB, $k$-NN, and SVM, have performed significantly worse than the control.
\item When using random subspace, DT has performed better than other classifiers, i.e. DS, NB, SVM, $k$-NN, OneR, CR, and BN.
\item When using nested dichotomies, the Rom test indicates that the control classifier is better than DS, NB, SVM, CR, and BN. 
\item When using dagging,  DS, CR, and SVM have unsatisfactory performance in comparison with the control classifier.

\end{itemize}

To sum up, the statistical tests have confirmed that the worst performers in all schemes are SVM, DS, and CR. Furthermore, based on the results we suggest to avoid RT, BN, and and NB for the next event prediction task. Overall, this analysis has highlighted that ensemble schemes in the next event prediction task of predictive monitoring are advantageous only when considered for badly performing classifiers. For high performing classifiers, such as DT, C-DT, and DTNB, the performance of individual classifiers is comparable to the one of ensemble schemes. It can be concluded, therefore, that ensemble learning is not a particularly high rewarding choice in the case of predicting the next event in a case in an event log. Users may be better off focusing on individual high performing classifiers, such as DT and C-DT.

\subsection{Threats to Validity}

As far as internal validity is concerned, the benchmark developed in this paper suffers from an intrinsic limitation related to the multiple degrees of freedom available while designing the experiments. For instance, more options for encoding features or other hyperparameter configurations for classifiers may have been considered. To keep the number of experiments and statistical tests manageable, however, we have decided to focus only on one type of encoding (window-based) that is used by all other approaches in the literature for the considered prediction task. Regarding hyperparameter configuration, our choice has been to avoid complex hyperparameter optimization, so as to develop models that can be quickly developed and easily managed even by practitioners with limited technical knowledge.

Furthermore, this work focuses mainly on evaluating the \emph{ability} of base learners in ensembles, but it does not extensively target their \emph{diversity}. While diversity is addressed by ensemble schemes that  train base classifiers on different samples, such as bagging, future work should consider different structures and parameter settings to increase diversity in traditional ensemble schemes~\cite{yin2014novel}. 

Regarding external validity, the generalizability of the results presented in this paper is restricted to the domain of predictive monitoring with event logs. As acknowledged by the recent publication of several benchmarks for predictive monitoring, there is a growing need for empirical studies that can support practitioners, who often lack deep technical machine learning knowledge, to choose the best model for their prediction task. The proposed benchmark fits within this line of applied research and the presented results should not be generalized to other domains.    

\section{Conclusions}
\label{conc}
This paper extends a benchmark previously published by the authors~\cite{tama2019empirical} regarding the next event prediction task in business process monitoring, by considering the effect of increasing the window size for encoding features and benchmarking the performance of 12 individual classification algorithms and 6 ensemble schemes. The benchmark has identified a set of high performing tree-based classifiers, the performance of which improves only slightly when considered in ensembles as opposed to being used as individual classifiers. Moreover, the benchmark highlights that ensemble schemes improve accuracy only in the case of low performing classifiers, such as SVM. Regarding the size of the window considered for encoding features, this benchmark did not obtain conclusive evidence. Generally, we suggest to assess the optimal window size for each event log, while we also highlight that lower window sizes, e.g., $l=3$ as consistently considered previously in the literature, strike a good balance between performance and availability of samples for training and testing. 

The proposed benchmark fills a gap in the literature related with providing informed guidelines about how to select a high performing machine learning model for given prediction tasks in business process predictive monitoring. Future work will concern extending this type of benchmark to different prediction tasks, such as outcome-based process prediction or prediction of remaining times in process execution. The benchmark can also be extended by considering other approaches for classification based on neural networks.  Also, the runtime performance of different classifiers will be benchmarked, which is particularly useful when choosing a classifier in scenarios in which predictions are supposed to be made in (near) real-time to support pro-active decision making scenarios.


\bibliographystyle{ACM-Reference-Format}
\bibliography{sample-base}

\newpage

\begin{appendices}

\section{List of Hyperparameter Values}
\label{param}
We provide here the list of hyperparameter settings used for each base classifier. Note that, when a parameter is not mentioned below, the default value in the considered implementation has been used.
\begin{itemize}
    \item Decision tree (J48)\\
    Confidence factor: 0.25; minimum number of instances per leaf: 2; number of folds used for reduced-error pruning: 3; pruning is performed: FALSE; reduced-error pruning: FALSE.
    \item Credal decision tree\\
    Parameter in imprecise Dirichlet model: 1.0; maximum tree depth: -1; minimum weight of the instances in a leaf: 2.0; no pruning: FALSE; amount of fold used for pruning: 3.
    \item Random tree\\
    Number of randomly chosen attributes: 0; allow unclassified instances: FALSE; break ties randomly: FALSE; maximum depth of the tree: unlimited; minimum weight of the instances in a leaf: 1.0; amount of data used for backfitting: 0.
    \item Decision stump\\
    Learning parameters are not available.
    \item Naive Bayes\\
    Use kernel estimator: FALSE.
    \item Support vector machine\\
    Type: $L2$-loss support vector machines (dual); bias term: 1.0; cost $C$: 1.0; termination criterion $\epsilon$: 0.01.
    \item $k$-Nearest neighbor\\
    Number of neighbor used: 2; cross-validation is used: FALSE; distance weighting: no; search algorithm: linear search.
    \item JRip\\
    Amount of folds used for pruning: 3; minimum weight of the instances in a rule: 2; number of optimization runs: 2, use pruning: TRUE.
    \item OneR\\
    Minimum bucket size: 6.
    \item Conjunctive rule\\
    Amount of folds used for pruning: 3; minimum weight of the instances in a rule: 2; number of antecedents allowed: -1.
    \item Bayesian network\\
    Estimator: simple; search algorithm: $K_{2}$; use the data structure for increasing speed: FALSE.
    \item DTNB\\
    Number of folds for cross-validation: 1 (leave one out); search algorithm: backwards with delete; use $k-NN$ instead of majority class: FALSE.
    
\end{itemize} 

\vspace{40pt}
\section{Performance Results}
\label{perf}
This section presents the detailed performance results of all classifier schemes over different event log datasets and window sizes.

\begin{table}
\caption{Results of average accuracy for each classifier and dataset as an individual classifier}
\label{tab5}
\centering
\resizebox{1.05\textwidth}{!}{
\begin{tabular}{llc|llllllllllll}
\hline
Variability& dataset&Window size& DT& C-DT& RT& DS& NB& SVM& k-NN  & JRip  & OneR & CR& BN& DTNB  \\\hline\hline
\multicolumn{1}{c}{\multirow{18}{*}{Low}} & \multirow{6}{*}{Helpdesk}&3& 66.42&	64.92&	58.41&	62.94&	63.22&	19.21&	58.41&	64.03&	65.04&	62.94&	63.59&	65.84 \\
&&4&73.99&	73.91&	65.82&	72.09&	70.03&	27.84&	65.88&	73.27&	74.12&	72.09&	69.98&	73.27\\
&&5&74.86&	75.34&	69.02&	72.87&	70.64&	26.98&	68.59&	72.87&	75.70&	72.87&	70.23&	73.05\\
&&6&74.74&	74.62&	71.11&	73.97&	74.61&	35.77&	72.40&	73.71&	77.20&	73.97&	74.48&	75.26\\
&&7&77.89&	72.92&	70.87&	68.55&	71.17&	35.24&	68.55&	74.71&	74.10&	69.73&	72.34&	75.85\\
&&8&76.45&	64.46&	69.07&	64.30&	73.29&	20.56&	70.65&	71.76&	77.56&	60.84&	77.48&	72.76\\\cline{2-15}

& \multirow{6}{*}{Hospital Billing} &3 & 92.63 & 92.58 & 88.10 & 61.43 & 90.98 & 59.07 & 88.10 & 92.61 & 92.27 & 61.43 & 90.93 & 92.51 \\
&&4 & 91.23 & 91.20 & 86.36 & 73.42 & 87.97 & 47.80 & 86.34 & 90.96 & 90.56 & 73.42 & 88.07 & 91.01 \\
&&5 & 88.31 & 88.33 & 83.05 & 58.26 & 83.84 & 30.60 & 82.97 & 87.91 & 87.45 & 58.26 & 83.95 & 87.76 \\
&&6 & 85.89 & 85.77 & 81.96 & 44.31 & 79.56 & 26.67 & 82.04 & 85.11 & 83.98 & 44.31 & 79.88 & 84.37 \\
&&7 & 85.41 & 85.05 & 81.03 & 48.94 & 77.63 & 30.17 & 80.71 & 84.68 & 84.27 & 48.94 & 78.19 & 84.03 \\
&&8 & 88.29 & 87.76 & 84.13 & 54.93 & 79.04 & 26.41 & 83.42 & 87.91 & 86.57 & 54.93 & 79.53 & 86.70\\\cline{2-15}

& \multirow{6}{*}{Road Traffic}     &3& 96.21 & 95.70 & 96.10 & 77.65 & 88.94 & 3.01  & 96.17 & 95.86 & 87.40 & 77.65 & 92.52 & 94.97 \\
&&4 & 88.68 & 88.39 & 87.31 & 79.10 & 85.38 & 19.89 & 87.39 & 88.82 & 81.03 & 79.01 & 85.89 & 87.34 \\
&&5 & 90.18 & 89.11 & 88.42 & 60.91 & 77.37 & 28.99 & 88.99 & 88.46 & 74.91 & 60.91 & 81.93 & 87.23 \\
&&6 & 85.27 & 84.18 & 86.68 & 54.32 & 80.21 & 14.26 & 87.21 & 82.70 & 75.90 & 54.67 & 77.52 & 84.18 \\
&&7 & 80.65 & 81.81 & 82.73 & 63.63 & 82.93 & 28.13 & 84.78 & 81.15 & 62.45 & 63.92 & 78.57 & 80.92 \\
&&8 & 69.49 & 71.87 & 74.85 & 78.75 & 78.05 & 42.10 & 74.23 & 75.48 & 79.89 & 77.54 & 73.79 & 79.34\\\hline

\multicolumn{3}{l|}{Average Friedman rank} &\textbf{2.50}& 	\underline{3.81}& 	7.11 &	9.28 &	6.83 &	12.00 &	6.92 &	4.25 &	4.89 &	9.31& 	7.06 &	\textit{4.06} 
  \\\hline

\multicolumn{1}{c}{\multirow{18}{*}{High}}                 & \multirow{6}{*}{BPIC2012}&
3 & 75.87 & 75.73 & 66.29 & 73.44 & 67.60 & 32.73 & 66.36 & 75.46 & 73.65 & 73.44 & 67.86 & 75.74 \\
&&4 & 76.56 & 76.45 & 69.21 & 76.18 & 76.40 & 28.13 & 69.28 & 76.60 & 76.40 & 75.97 & 76.44 & 76.46 \\
&&5 & 71.26 & 71.25 & 64.26 & 71.19 & 71.29 & 28.34 & 64.18 & 71.06 & 71.19 & 71.19 & 71.31 & 71.22 \\
&&6 & 70.51 & 69.42 & 60.22 & 69.49 & 69.27 & 41.95 & 60.44 & 68.10 & 65.94 & 69.27 & 69.12 & 70.75 \\
&&7 & 67.42 & 63.84 & 59.60 & 64.10 & 63.83 & 48.91 & 59.60 & 65.03 & 59.77 & 63.24 & 63.11 & 67.16 \\
&&8 & 69.73 & 67.51 & 57.69 & 67.06 & 62.01 & 40.54 & 57.24 & 70.21 & 62.56 & 66.33 & 61.27 & 71.69
\\\cline{2-15}

& \multirow{6}{*}{BPIC2013 Incident} & 
3 & 64.09 & 64.80 & 59.35 & 50.17 & 60.80 & 34.06 & 59.32 & 52.19 & 61.70 & 50.17 & 62.77 & 63.96 \\
&&4 & 61.20 & 61.83 & 55.14 & 46.15 & 57.38 & 26.97 & 55.12 & 47.81 & 58.41 & 46.09 & 58.86 & 61.02 \\
&&5 & 62.97 & 63.72 & 57.53 & 49.20 & 59.13 & 22.37 & 57.32 & 50.89 & 60.22 & 49.20 & 60.80 & 62.95 \\
&&6 & 61.39 & 62.35 & 55.48 & 47.30 & 57.32 & 27.18 & 55.40 & 49.71 & 58.94 & 47.30 & 59.58 & 61.64 \\
&&7 & 62.18 & 62.94 & 56.62 & 48.66 & 58.36 & 26.15 & 55.31 & 51.61 & 59.81 & 48.66 & 59.60 & 62.42 \\
&&8 & 61.47 & 62.22 & 55.16 & 47.31 & 57.52 & 30.08 & 53.55 & 50.97 & 59.16 & 47.45 & 59.03 & 61.69
\\\cline{2-15}

& \multirow{5}{*}{Sepsis}&
3 & 53.32 & 53.69 & 45.83 & 29.41 & 48.77 & 24.82 & 45.72 & 49.60 & 48.83 & 29.66 & 49.12 & 53.52 \\
&&4 & 58.81 & 59.09 & 51.40 & 29.27 & 51.54 & 22.76 & 51.17 & 56.74 & 50.15 & 29.27 & 51.55 & 57.40 \\
&&5 & 59.34 & 59.14 & 52.10 & 29.50 & 52.32 & 19.63 & 52.09 & 57.44 & 49.88 & 29.63 & 52.56 & 57.08 \\
&&6 & 58.47 & 58.31 & 50.06 & 30.06 & 54.53 & 21.95 & 49.20 & 56.78 & 51.45 & 30.19 & 54.81 & 55.98 \\
&&7 & 57.77 & 57.85 & 49.74 & 34.29 & 52.98 & 21.34 & 48.09 & 56.19 & 54.12 & 34.29 & 53.21 & 55.24 \\
&&8 & 54.72 & 55.02 & 46.05 & 35.29 & 50.94 & 22.96 & 43.10 & 53.93 & 53.45 & 35.79 & 50.75 & 52.96
\\\hline
\multicolumn{3}{l|}{Average Friedman rank} &  \textbf{2.11}& 	\textbf{2.11}& 	8.42& 	8.97& 	6.17& 	12.00& 	9.03& 	5.61& 	6.19& 	9.17& 	\textit{5.39}& 	\underline{2.83} 
  \\\hline
\end{tabular}
}
\end{table}

\begin{table}
\caption{Results of average accuracy for each classifier and dataset as a base classifier in Bagging}
\label{tab6}
\centering
\resizebox{1.05\textwidth}{!}{
\begin{tabular}{llc|llllllllllll}
\hline
Variability& dataset&Window size& DT& C-DT& RT& DS& NB& SVM& k-NN  & JRip  & OneR & CR& BN& DTNB\\\hline\hline
\multirow{18}{*}{Low}            & \multirow{6}{*}{Helpdesk}&
3 & 65.38 & 65.88 & 59.69 & 62.94 & 63.36 & 25.63 & 58.64 & 65.48 & 55.63 & 62.47 & 63.57 & 66.7  \\
&&4 & 74.12 & 73.77 & 67.48 & 72.09 & 69.51 & 18.49 & 66.06 & 72.88 & 73.99 & 72.09 & 70.16 & 73.59 \\
&&5 & 74.37 & 75.52 & 71.48 & 72.87 & 71.19 & 39.40 & 69.25 & 73.95 & 75.52 & 72.87 & 70.59 & 75.82 \\
&&6 & 77.58 & 76.03 & 75.39 & 73.44 & 74.61 & 47.42 & 72.66 & 75.39 & 77.20 & 73.58 & 73.83 & 77.34 \\
&&7 & 78.76 & 76.13 & 75.85 & 69.42 & 72.07 & 38.71 & 69.13 & 78.13 & 74.66 & 68.55 & 72.93 & 78.46 \\
&&8 & 79.67 & 75.95 & 76.37 & 68.63 & 70.04 & 30.89 & 70.15 & 70.07 & 76.48 & 66.05 & 74.26 & 73.90
\\\cline{2-15}

& \multirow{6}{*}{Hospital Billing} &
3 & 92.5  & 92.63 & 88.91 & 61.43 & 91.01 & 69.29 & 88.73 & 92.43 & 92.27 & 61.43 & 90.99 & 92.51 \\
&&4 & 91.23 & 91.29 & 87.30 & 73.42 & 87.97 & 65.80 & 86.91 & 90.91 & 90.57 & 73.42 & 88.06 & 91.15 \\
&&5 & 88.15 & 88.33 & 84.05 & 58.26 & 83.89 & 54.26 & 83.52 & 87.93 & 87.46 & 58.26 & 83.89 & 87.89 \\
&&6 & 85.94 & 85.79 & 83.24 & 44.31 & 79.60 & 40.51 & 82.37 & 85.52 & 83.83 & 44.31 & 80.05 & 85.21 \\
&&7 & 85.51 & 85.43 & 82.60 & 48.94 & 77.87 & 43.11 & 80.94 & 85.32 & 84.22 & 48.94 & 78.29 & 84.44 \\
&&8 & 88.43 & 87.90 & 86.10 & 54.93 & 79.43 & 47.01 & 83.60 & 88.09 & 86.55 & 54.93 & 79.80 & 86.59
\\\cline{2-15}

& \multirow{6}{*}{Road Traffic}     &
3 & 96.27 & 95.89 & 96.16 & 77.65 & 88.95 & 18.54 & 96.13 & 96.00 & 87.40  & 77.65 & 92.56 & 95.44 \\
&&4 & 88.56 & 88.41 & 87.89 & 79.05 & 85.35 & 23.48 & 87.83 & 88.59 & 81.04 & 79.08 & 85.86 & 87.26 \\
&&5 & 89.99 & 88.84 & 88.72 & 60.91 & 78.33 & 18.98 & 88.30 & 89.15 & 74.80 & 61.03 & 82.01 & 87.76 \\
&&6 & 87.60 & 85.80 & 87.04 & 63.13 & 79.47 & 25.40 & 86.51 & 83.61 & 74.11 & 63.47 & 77.15 & 85.61 \\
&&7 & 84.40 & 82.95 & 84.46 & 66.54 & 84.12 & 37.96 & 84.77 & 82.35 & 64.82 & 67.44 & 79.46 & 83.85 \\
&&8 & 69.45 & 76.84 & 73.64 & 77.50 & 79.93 & 48.82 & 71.80 & 76.76 & 81.14 & 78.12 & 70.70 & 79.37
\\\hline
\multicolumn{3}{l|}{Average Friedman rank} &  \textbf{2.28}& 	\underline{3.14}&	5.69 &	9.53 &	7.36 &	11.89&	7.44 &	4.19 &	5.81 &	9.42 &	7.47 &	\textit{3.78} 
  \\\hline

\multirow{18}{*}{High}           & \multirow{6}{*}{BPIC2012}          &
3 & 75.91 & 75.88 & 67.07 & 73.44 & 67.59 & 45.73 & 67.01 & 75.57 & 73.64 & 73.44 & 67.90 & 75.71 \\
&&4 & 76.47 & 76.54 & 69.52 & 76.18 & 76.44 & 59.77 & 69.82 & 76.19 & 73.98 & 76.18 & 76.44 & 76.47 \\
&&5 & 71.13 & 71.20 & 64.49 & 71.09 & 71.29 & 41.28 & 65.00 & 69.89 & 56.03 & 71.11 & 71.31 & 71.31 \\
&&6 & 69.70 & 69.73 & 60.38 & 69.24 & 69.39 & 63.47 & 60.50 & 68.44 & 56.67 & 68.78 & 69.12 & 71.06 \\
&&7 & 65.57 & 65.37 & 59.94 & 64.76 & 64.17 & 57.40 & 60.53 & 64.37 & 58.20 & 64.96 & 63.11 & 67.49 \\
&&8 & 64.98 & 65.45 & 58.14 & 67.52 & 62.76 & 60.20 & 57.98 & 66.94 & 60.94 & 66.63 & 61.87 & 71.38
\\\cline{2-15}

& \multirow{6}{*}{BPIC2013 Incident} &
3 & 62.14 & 64.30 & 59.47 & 50.17 & 61.09 & 39.58 & 59.31 & 52.89 & 61.69 & 50.17 & 62.88 & 63.86 \\
&&4 & 58.61 & 61.32 & 55.50 & 49.75 & 57.46 & 36.42 & 55.20 & 48.83 & 58.38 & 50.94 & 58.90 & 61.31 \\
&&5 & 60.83 & 63.45 & 58.18 & 49.20 & 58.93 & 33.67 & 57.51 & 51.94 & 60.22 & 49.20 & 60.81 & 63.48 \\
&&6 & 59.63 & 61.89 & 57.06 & 47.30 & 57.36 & 31.95 & 55.74 & 50.98 & 58.94 & 47.30 & 59.58 & 60.53 \\
&&7 & 60.25 & 62.72 & 57.76 & 48.66 & 58.64 & 39.21 & 55.72 & 53.06 & 59.82 & 48.66 & 59.60 & 60.53 \\
&&8 & 59.41 & 62.12 & 57.06 & 50.96 & 57.52 & 37.67 & 53.98 & 52.35 & 59.15 & 48.68 & 59.06 & 59.99
\\\cline{2-15}

& \multirow{6}{*}{Sepsis}&
3 & 50.83 & 53.32 & 46.6  & 31.25 & 48.67 & 31.21 & 46.34 & 51.66 & 48.8  & 33.39 & 49.07 & 53.92 \\
&&4 & 56.81 & 59.51 & 52.46 & 31.91 & 51.03 & 28.18 & 51.81 & 58.69 & 50.07 & 31.89 & 51.45 & 58.10 \\
&&5 & 58.14 & 60.00 & 53.98 & 30.30 & 52.25 & 23.95 & 52.55 & 60.01 & 49.78 & 30.71 & 52.59 & 58.41 \\
&&6 & 56.97 & 58.84 & 53.18 & 34.23 & 54.36 & 25.38 & 49.57 & 58.97 & 51.35 & 33.47 & 54.80 & 57.61 \\
&&7 & 56.97 & 58.21 & 53.50 & 34.29 & 53.09 & 25.69 & 48.36 & 57.36 & 53.97 & 34.29 & 53.21 & 56.90 \\
&&8 & 54.55 & 55.96 & 50.56 & 35.96 & 50.94 & 29.83 & 43.28 & 54.95 & 53.42 & 37.80 & 50.94 & 53.14
\\\hline
\multicolumn{3}{l|}{Average Friedman rank}&\textit{3.42} &	\textbf{1.83} &	8.00 &	8.81 &	6.33 &	11.72 &	8.78 &	5.67 &	7.00 &	8.86& 	5.36& 	\underline{2.22} 
 \\\hline
\end{tabular}
}
\end{table}

\begin{table}
\caption{Results of average accuracy for each classifier and dataset as a base classifier in Boosting}
\label{tab7}
\centering
\resizebox{1.05\textwidth}{!}{
\begin{tabular}{llc|llllllllllll}
\hline
Variability& dataset&Window size& DT& C-DT& RT& DS& NB& SVM& k-NN  & JRip  & OneR & CR& BN& DTNB\\\hline\hline
\multirow{18}{*}{Low}            & \multirow{6}{*}{Helpdesk}         &
3 & 66.26 & 65.15 & 58.66 & 62.94 & 63.22 & 23.98 & 58.27 & 64.44 & 61.49 & 62.94 & 63.59 & 66.47 \\
&&4 & 72.27 & 74.07 & 66.14 & 72.09 & 70.03 & 21.24 & 65.80 & 73.01 & 72.30 & 72.09 & 69.98 & 73.14 \\
&&5 & 72.98 & 75.63 & 69.14 & 72.87 & 70.64 & 40.73 & 68.59 & 73.29 & 74.49 & 72.87 & 70.23 & 74.24 \\
&&6 & 71.77 & 74.87 & 70.98 & 73.97 & 74.62 & 36.63 & 72.40 & 74.08 & 76.15 & 73.97 & 74.48 & 76.15 \\
&&7 & 72.92 & 74.37 & 71.18 & 68.55 & 69.45 & 26.28 & 68.55 & 76.74 & 73.22 & 69.73 & 72.67 & 79.34 \\
&&8 & 73.87 & 70.66 & 72.74 & 64.30 & 68.57 & 26.71 & 70.65 & 71.59 & 73.84 & 64.42 & 71.77 & 68.55
\\\cline{2-15}

& \multirow{6}{*}{Hospital Billing} &
3 & 88.53 & 92.61 & 88.09 & 61.43 & 90.98 & 47.71 & 88.09 & 92.61 & 92.24 & 61.43 & 90.93 & 92.51 \\
&&4 & 87.31 & 91.21 & 86.40 & 73.42 & 87.97 & 49.13 & 86.32 & 90.96 & 90.57 & 73.42 & 88.07 & 91.01 \\
&&5 & 86.81 & 88.30 & 83.10 & 58.26 & 83.84 & 38.81 & 82.92 & 87.90 & 87.46 & 58.26 & 83.95 & 87.76 \\
&&6 & 83.78 & 85.92 & 82.53 & 44.31 & 79.56 & 25.71 & 81.97 & 84.56 & 83.49 & 44.31 & 79.88 & 85.54 \\
&&7 & 82.76 & 85.28 & 81.65 & 48.94 & 77.63 & 28.10 & 80.67 & 84.43 & 84.22 & 48.94 & 78.19 & 85.06 \\
&&8 & 85.19 & 87.91 & 84.57 & 54.93 & 81.27 & 31.39 & 83.36 & 87.40 & 86.46 & 54.93 & 81.80 & 87.61
\\\cline{2-15}

& \multirow{6}{*}{Road Traffic}     &
3 & 96.27 & 95.88 & 96.14 & 77.65 & 89.13 & 11.66 & 96.17 & 95.86 & 87.08 & 77.65 & 94.86 & 95.46 \\
&&4 & 88.48 & 88.53 & 87.62 & 79.10 & 85.81 & 15.59 & 87.94 & 88.52 & 80.50 & 79.01 & 87.32 & 88.64 \\
&&5 & 88.65 & 90.14 & 88.80 & 60.91 & 83.89 & 26.71 & 89.03 & 89.15 & 77.18 & 60.91 & 86.11 & 88.92 \\
&&6 & 86.72 & 85.22 & 85.94 & 54.32 & 81.09 & 20.23 & 87.40 & 87.78 & 74.60 & 54.67 & 84.16 & 88.14 \\
&&7 & 83.30 & 82.96 & 82.73 & 63.63 & 80.34 & 28.83 & 84.78 & 84.08 & 72.10 & 63.92 & 81.26 & 86.22 \\
&&8 & 80.48 & 79.26 & 78.53 & 78.75 & 75.00 & 34.45 & 74.23 & 72.16 & 83.01 & 77.54 & 76.87 & 78.12
\\\hline
\multicolumn{3}{l|}{Average Friedman rank} &  4.56& 	\textbf{2.47}& 	6.86 &	9.42 &	7.61 &	12.00 &	7.33 &	\textit{3.64} &	5.19 &	9.33 &	6.78& 	\underline{2.81} 
  \\\hline

\multirow{18}{*}{High}           & \multirow{6}{*}{BPIC2012}          &
3 & 75.87 & 75.67 & 66.30 & 73.44 & 67.60 & 38.45 & 66.28 & 75.43 & 69.83 & 73.44 & 67.86 & 75.74 \\
&&4 & 76.41 & 76.45 & 69.13 & 76.18 & 76.40 & 39.05 & 69.37 & 76.40 & 67.39 & 75.97 & 76.44 & 76.46 \\
&&5 & 71.36 & 71.33 & 64.21 & 71.19 & 71.31 & 40.78 & 64.15 & 70.60 & 64.05 & 71.19 & 71.17 & 71.31 \\
&&6 & 70.54 & 69.09 & 60.22 & 69.49 & 70.04 & 45.32 & 60.38 & 68.62 & 61.95 & 69.27 & 70.54 & 70.88 \\
&&7 & 67.56 & 62.65 & 59.47 & 63.97 & 65.09 & 34.15 & 59.60 & 64.77 & 61.31 & 63.38 & 66.56 & 67.42 \\
&&8 & 71.68 & 64.70 & 57.24 & 67.06 & 69.46 & 49.40 & 57.24 & 65.75 & 60.51 & 63.95 & 70.35 & 72.73
\\\cline{2-15}

& \multirow{6}{*}{BPIC2013 Incident} &
3 & 62.89 & 64.74 & 59.34 & 50.17 & 60.80 & 29.99 & 59.32 & 52.19 & 61.65 & 50.17 & 62.77 & 63.96 \\
&&4 & 59.38 & 61.93 & 55.16 & 46.15 & 57.38 & 29.15 & 55.12 & 47.81 & 58.37 & 46.09 & 58.86 & 61.02 \\
&&5 & 60.43 & 63.84 & 57.45 & 49.20 & 59.13 & 20.80 & 57.04 & 51.03 & 60.21 & 49.20 & 60.80 & 62.95 \\
&&6 & 58.66 & 62.44 & 56.14 & 47.30 & 57.32 & 15.98 & 55.28 & 49.84 & 58.94 & 47.30 & 59.58 & 61.64 \\
&&7 & 59.18 & 62.97 & 56.92 & 48.66 & 58.36 & 24.18 & 55.30 & 51.61 & 59.80 & 48.66 & 59.60 & 62.42 \\
&&8 & 57.71 & 62.28 & 55.84 & 47.31 & 57.52 & 31.59 & 53.54 & 50.97 & 59.14 & 47.45 & 59.03 & 61.69
\\\cline{2-15}

& \multirow{6}{*}{Sepsis}           &
3 & 53.10 & 53.73 & 45.82 & 29.41 & 48.77 & 24.62 & 45.71 & 49.66 & 48.76 & 29.66 & 49.12 & 53.51 \\
&&4 & 57.84 & 58.90 & 51.66 & 29.27 & 51.67 & 21.37 & 51.17 & 56.74 & 49.93 & 29.27 & 51.87 & 57.48 \\
&&5 & 57.43 & 59.24 & 52.52 & 29.50 & 52.32 & 20.31 & 52.07 & 57.44 & 49.71 & 29.63 & 52.56 & 58.03 \\
&&6 & 55.56 & 58.35 & 51.15 & 30.06 & 54.53 & 18.49 & 49.22 & 56.78 & 51.21 & 30.19 & 54.81 & 56.57 \\
&&7 & 55.44 & 57.58 & 51.83 & 34.29 & 52.98 & 19.70 & 48.17 & 56.19 & 53.99 & 34.29 & 53.21 & 55.82 \\
&&8 & 51.85 & 55.56 & 48.19 & 35.29 & 50.94 & 22.04 & 43.09 & 53.93 & 53.31 & 35.79 & 50.75 & 52.96
\\\hline
\multicolumn{3}{l|}{Average Friedman rank} &  \textit{3.25}& 	\underline{2.28} &	8.25 &	9.00 &	5.78&	12.00 &	8.92 &	5.97& 	6.72 &	9.11 &	4.53 &	\textbf{2.19} 
  \\\hline
\end{tabular}
}
\end{table}

\begin{table}
\caption{Results of average accuracy for each classifier and dataset as a base classifier in Random Subspace}
\label{tab9}
\centering
\resizebox{1.05\textwidth}{!}{
\begin{tabular}{llc|llllllllllll}
\hline
Variability& dataset&Window size& DT& C-DT& RT& DS& NB& SVM& k-NN  & JRip  & OneR & CR& BN& DTNB\\\hline\hline
\multirow{18}{*}{Low}            & \multirow{6}{*}{Helpdesk}         &
3 & 64.83 & 64.27 & 59.14 & 60.56 & 64.27 & 42.95 & 59.18 & 61.96 & 62.14 & 59.31 & 64.03 & 64.29 \\
&&4 & 73.17 & 73.54 & 66.28 & 70.67 & 70.85 & 33.54 & 66.17 & 72.51 & 72.22 & 70.67 & 70.96 & 72.72 \\
&&5 & 74.67 & 75.33 & 70.76 & 71.48 & 72.57 & 55.65 & 70.28 & 74.37 & 74.49 & 71.54 & 72.63 & 74.31 \\
&&6 & 76.55 & 76.55 & 73.20 & 74.10 & 73.96 & 50.67 & 73.71 & 75.90 & 77.20 & 73.71 & 74.73 & 75.27 \\
&&7 & 76.72 & 73.51 & 75.81 & 69.14 & 70.61 & 58.58 & 71.72 & 74.39 & 73.81 & 69.72 & 70.87 & 72.62 \\
&&8 & 76.45 & 75.95 & 78.06 & 65.96 & 71.20 & 52.17 & 73.26 & 70.63 & 74.92 & 62.83 & 73.35 & 73.90 
\\\cline{2-15}

& \multirow{6}{*}{Hospital Billing} &
3 & 90.99 & 91.11 & 87.44 & 63.28 & 90.36 & 67.78 & 87.43 & 88.67 & 89.28 & 63.28 & 90.40 & 90.89 \\
&&4 & 90.65 & 90.68 & 87.42 & 73.41 & 88.38 & 68.75 & 87.43 & 89.36 & 88.80 & 73.41 & 88.47 & 90.04 \\
&&5 & 86.95 & 87.15 & 83.03 & 61.05 & 83.57 & 58.84 & 83.05 & 82.13 & 81.84 & 61.05 & 83.69 & 86.55 \\
&&6 & 84.42 & 84.61 & 79.84 & 44.31 & 80.07 & 55.92 & 79.84 & 82.43 & 77.62 & 44.31 & 80.13 & 83.68 \\
&&7 & 84.33 & 84.75 & 81.02 & 48.94 & 77.75 & 57.66 & 80.57 & 83.65 & 80.79 & 48.94 & 78.27 & 83.74 \\
&&8 & 88.17 & 87.90 & 86.49 & 54.93 & 81.81 & 58.52 & 85.60 & 87.79 & 85.74 & 54.93 & 82.10 & 87.00 
\\\cline{2-15}

& \multirow{6}{*}{Road Traffic}     &
3 & 92.76 & 92.24 & 92.82 & 77.65 & 85.62 & 64.12 & 92.84 & 90.18 & 84.38 & 77.65 & 89.71 & 91.1  \\
&&4 & 86.24 & 85.84 & 85.82 & 75.64 & 81.98 & 59.84 & 85.86 & 85.20 & 78.73 & 75.58 & 82.27 & 84.97 \\
&&5 & 84.88 & 82.38 & 84.66 & 61.68 & 74.56 & 59.31 & 84.62 & 82.69 & 70.61 & 61.91 & 75.59 & 81.08 \\
&&6 & 84.56 & 83.47 & 86.51 & 65.09 & 78.78 & 44.25 & 85.96 & 83.67 & 70.86 & 66.15 & 77.56 & 81.15 \\
&&7 & 79.47 & 80.95 & 83.01 & 67.16 & 80.29 & 58.59 & 84.17 & 79.75 & 64.50 & 67.16 & 79.76 & 77.45 \\
&&8 & 71.21 & 74.96 & 76.73 & 80.59 & 76.76 & 51.88 & 72.43 & 75.51 & 72.43 & 76.91 & 74.34 & 81.14
\\\hline

\multicolumn{3}{l|}{Average Friedman rank} &  \textbf{2.64}& 	\underline{2.78} &	5.69 &	9.69 &	6.81 &	11.56 &	6.58 &	5.11 &	6.69 &	9.83 &	6.33 &	\textit{4.28} 
 \\\hline

\multirow{18}{*}{High}           & \multirow{6}{*}{BPIC2012}          &
3 & 75.59 & 75.57 & 68.25 & 73.44 & 73.46 & 48.82 & 68.30 & 74.38 & 73.66 & 71.49 & 72.77 & 75.44 \\
&&4 & 76.57 & 76.55 & 70.45 & 76.19 & 76.36 & 56.61 & 70.52 & 76.41 & 76.37 & 76.20 & 76.47 & 76.49 \\
&&5 & 71.22 & 71.42 & 67.26 & 71.09 & 71.29 & 70.02 & 67.23 & 71.54 & 71.09 & 71.09 & 71.31 & 71.31 \\
&&6 & 69.89 & 69.64 & 61.80 & 69.33 & 69.36 & 60.60 & 61.83 & 68.81 & 65.54 & 69.21 & 69.18 & 70.04 \\
&&7 & 66.22 & 63.17 & 60.92 & 65.23 & 63.77 & 49.36 & 61.18 & 65.44 & 61.89 & 64.10 & 63.57 & 66.57 \\
&&8 & 67.97 & 67.51 & 59.47 & 67.22 & 62.90 & 53.62 & 59.91 & 70.19 & 64.05 & 67.81 & 62.62 & 70.51 
\\\cline{2-15}

& \multirow{6}{*}{BPIC2013 Incident} &
3 & 59.79 & 61.02 & 55.49 & 48.45 & 59.06 & 41.79 & 55.58 & 48.93 & 54.06 & 48.66 & 60.44 & 60.94 \\
&&4 & 60.22 & 60.82 & 55.65 & 46.19 & 58.81 & 40.56 & 55.69 & 46.41 & 57.48 & 46.87 & 59.63 & 60.45 \\
&&5 & 61.73 & 62.17 & 58.40 & 49.20 & 61.10 & 41.67 & 58.43 & 49.15 & 58.77 & 49.20 & 61.19 & 61.94 \\
&&6 & 60.84 & 61.62 & 56.38 & 47.60 & 59.02 & 42.63 & 56.31 & 47.52 & 55.81 & 48.07 & 60.08 & 61.24 \\
&&7 & 60.71 & 61.42 & 56.10 & 49.59 & 58.85 & 40.04 & 56.12 & 48.56 & 54.20 & 49.93 & 59.57 & 61.03 \\
&&8 & 60.56 & 61.25 & 56.67 & 48.00 & 58.36 & 42.72 & 56.53 & 47.66 & 55.58 & 48.15 & 59.20 & 60.78
\\\cline{2-15}

& \multirow{6}{*}{Sepsis}           &
3 & 49.05 & 50.9  & 42.53 & 30.42 & 47.49 & 34.27 & 42.58 & 48.12 & 39.61 & 30.45 & 47.67 & 50.15 \\
&&4 & 56.09 & 57.50 & 47.79 & 29.21 & 51.45 & 32.44 & 47.85 & 52.88 & 48.87 & 29.19 & 51.79 & 56.21 \\
&&5 & 56.35 & 57.45 & 48.43 & 30.20 & 52.40 & 34.49 & 48.42 & 53.29 & 45.39 & 30.29 & 52.54 & 55.39 \\
&&6 & 57.01 & 57.29 & 51.30 & 31.38 & 54.32 & 36.11 & 50.98 & 56.12 & 47.16 & 31.66 & 54.54 & 56.13 \\
&&7 & 53.99 & 55.03 & 48.72 & 35.70 & 52.88 & 36.40 & 48.46 & 55.18 & 49.08 & 35.50 & 52.97 & 54.40 \\
&&8 & 54.67 & 54.87 & 48.92 & 41.13 & 52.90 & 38.12 & 48.15 & 54.23 & 48.05 & 40.77 & 52.97 & 53.91
\\\hline
\multicolumn{3}{l|}{Average Friedman rank}&\textit{2.78}& \textbf{1.94} &	8.56& 	9.36 &	5.67& 	11.33& 	8.33 &	5.89 &	7.67 &	9.08& 	5.08 &	\underline{2.31}\\\hline
\end{tabular}
}
\end{table}

\begin{table}
\caption{Results of average accuracy for each classifier and dataset as a base classifier in Nested Dichotomies}
\label{tab10}
\centering
\resizebox{1.05\textwidth}{!}{
\begin{tabular}{llc|llllllllllll}
\hline
Variability& dataset&Window size& DT& C-DT& RT& DS& NB& SVM& k-NN  & JRip  & OneR & CR& BN& DTNB\\\hline\hline
\multirow{18}{*}{Low}            & \multirow{6}{*}{Helpdesk}         &
3 & 65.75 & 65.85 & 58.53 & 62.94 & 63.31 & 34.58 & 58.45 & 65.31 & 62.62 & 62.47 & 63.25 & 66.04 \\
&&4 & 73.88 & 72.77 & 65.96 & 71.93 & 69.88 & 21.23 & 65.88 & 73.42 & 73.17 & 73.30 & 69.95 & 73.30 \\
&&5 & 74.86 & 74.67 & 69.20 & 72.93 & 69.68 & 50.64 & 68.59 & 74.62 & 74.74 & 72.75 & 69.50 & 72.33 \\
&&6 & 75.26 & 76.69 & 71.38 & 71.89 & 72.41 & 29.31 & 72.40 & 73.06 & 75.39 & 71.65 & 73.05 & 75.64 \\
&&7 & 78.18 & 72.33 & 70.03 & 69.72 & 69.15 & 36.96 & 68.55 & 77.31 & 76.46 & 69.72 & 70.92 & 77.05 \\
&&8 & 77.56 & 68.74 & 71.24 & 66.44 & 68.96 & 16.11 & 70.65 & 71.76 & 78.14 & 55.92 & 70.08 & 73.31 \\
\cline{2-15}

& \multirow{6}{*}{Hospital Billing} &
3 & 92.59 & 92.54 & 88.09 & 88.54 & 90.33 & 53.96 & 88.08 & 92.43 & 92.27 & 88.36 & 90.20 & 92.52 \\
&&4 & 91.12 & 91.12 & 86.34 & 87.04 & 86.09 & 51.08 & 86.32 & 91.07 & 90.57 & 87.54 & 86.60 & 90.82 \\
&&5 & 88.24 & 88.21 & 82.76 & 80.69 & 82.15 & 42.98 & 82.89 & 87.86 & 87.44 & 80.66 & 82.06 & 87.64 \\
&&6 & 85.55 & 85.68 & 81.66 & 71.80 & 72.95 & 33.58 & 81.99 & 85.35 & 83.82 & 70.83 & 74.48 & 84.69 \\
&&7 & 85.12 & 84.95 & 81.18 & 78.29 & 73.99 & 27.69 & 80.66 & 84.87 & 84.18 & 78.10 & 74.52 & 84.19 \\
&&8 & 87.78 & 88.08 & 83.83 & 78.58 & 71.27 & 32.20 & 83.35 & 87.64 & 86.56 & 78.94 & 71.24 & 86.08 \\
\cline{2-15}

& \multirow{6}{*}{Road Traffic}     &
3 & 96.11 & 95.91 & 96.13 & 83.83 & 87.34 & 21.83 & 96.23 & 95.86 & 90.4  & 83.94 & 90.58 & 95.37 \\
&&4 & 88.49 & 88.39 & 87.30 & 80.62 & 79.74 & 33.46 & 87.72 & 88.23 & 84.41 & 81.25 & 83.36 & 88.08 \\
&&5 & 89.15 & 88.41 & 88.42 & 72.85 & 76.56 & 12.29 & 88.92 & 88.72 & 79.60 & 76.15 & 79.28 & 87.61 \\
&&6 & 80.38 & 78.58 & 84.72 & 73.88 & 73.18 & 18.39 & 87.57 & 83.46 & 80.72 & 73.16 & 67.80 & 82.41 \\
&&7 & 79.42 & 75.64 & 82.14 & 72.04 & 74.73 & 35.24 & 84.78 & 82.40 & 72.38 & 66.80 & 70.31 & 78.88 \\
&&8 & 70.74 & 69.63 & 77.24 & 74.41 & 79.34 & 27.83 & 74.23 & 73.16 & 73.82 & 68.79 & 76.80 & 79.96\\
\hline

\multicolumn{3}{l|}{Average Friedman rank} &\textbf{2.47}& 	\textit{3.86}& 	6.56& 	8.47 &	8.28& 	12.00&	6.72& 	\underline{3.56}& 	5.22&	9.06 &	7.94& 	\textit{3.86} 
  \\\hline

\multirow{18}{*}{High}           & \multirow{6}{*}{BPIC2012}          &
3 & 75.85 & 75.41 & 66.24 & 73.60 & 70.10 & 39.13 & 66.33 & 75.43 & 74.37 & 73.58 & 70.19 & 75.68 \\
&&4 & 76.38 & 76.08 & 69.16 & 76.48 & 76.31 & 42.15 & 69.33 & 76.55 & 76.38 & 76.37 & 76.38 & 76.51 \\
&&5 & 70.89 & 69.82 & 64.18 & 71.36 & 71.31 & 46.52 & 64.17 & 70.97 & 71.28 & 71.22 & 71.31 & 71.28 \\
&&6 & 70.20 & 67.54 & 60.10 & 69.55 & 69.33 & 39.49 & 60.31 & 68.47 & 66.06 & 68.47 & 69.21 & 71.00 \\
&&7 & 67.10 & 65.56 & 59.47 & 64.50 & 63.84 & 48.21 & 59.60 & 65.24 & 58.66 & 63.24 & 63.70 & 67.09 \\
&&8 & 67.97 & 64.71 & 57.39 & 67.82 & 61.72 & 53.58 & 57.24 & 69.31 & 62.27 & 66.64 & 61.28 & 73.03 \\

\cline{2-15}

& \multirow{6}{*}{BPIC2013 Incident} &
3 & 63.87 & 64.23 & 59.32 & 57.71 & 60.23 & 35.31 & 59.01 & 61.90 & 58.38 & 56.13 & 61.31 & 63.76 \\
&&4 & 61.13 & 61.06 & 55.12 & 53.75 & 57.08 & 21.59 & 54.82 & 60.08 & 55.86 & 53.09 & 57.79 & 60.90 \\
&&5 & 61.76 & 62.97 & 57.22 & 57.29 & 58.03 & 30.27 & 57.21 & 60.71 & 56.90 & 55.88 & 58.78 & 62.01 \\
&&6 & 61.48 & 61.85 & 55.54 & 56.42 & 56.64 & 23.76 & 55.26 & 60.63 & 56.51 & 59.55 & 57.61 & 60.58 \\
&&7 & 61.61 & 62.52 & 56.12 & 55.50 & 56.20 & 29.58 & 55.26 & 61.45 & 57.96 & 55.62 & 57.16 & 60.17 \\
&&8 & 60.16 & 61.96 & 54.99 & 56.06 & 57.18 & 30.45 & 53.51 & 60.50 & 56.10 & 54.68 & 57.67 & 58.53 \\
\cline{2-15}

& \multirow{6}{*}{Sepsis}           &
3 & 53.65 & 53.35 & 45.7  & 45.36 & 48.15 & 22.23 & 45.75 & 51.25 & 47.68 & 44.81 & 48.47 & 53.51 \\
&&4 & 59.15 & 58.57 & 50.79 & 44.76 & 50.29 & 20.24 & 51.22 & 56.46 & 47.79 & 44.39 & 50.36 & 58.08 \\
&&5 & 59.67 & 59.23 & 51.79 & 47.40 & 51.37 & 16.05 & 52.15 & 55.99 & 45.55 & 44.65 & 51.68 & 57.14 \\
&&6 & 58.48 & 57.61 & 49.30 & 47.64 & 51.73 & 18.71 & 49.20 & 54.79 & 44.71 & 44.36 & 51.87 & 55.92 \\
&&7 & 57.35 & 56.83 & 49.31 & 50.85 & 49.61 & 20.49 & 48.13 & 54.95 & 51.39 & 49.11 & 49.33 & 54.70 \\
&&8 & 54.67 & 54.79 & 45.26 & 52.09 & 50.21 & 21.84 & 43.10 & 53.92 & 50.48 & 52.19 & 50.07 & 50.31\\
\hline
\multicolumn{3}{l|}{Average Friedman rank}&\textbf{2.22} &	\textit{3.22} &	8.89 &	6.94 &	6.58& 	12.00& 	9.33 &	3.64& 	7.47 &	8.58 &	6.08 &	\underline{3.03} 
  \\\hline
\end{tabular}
}
\end{table}

\begin{table}
\caption{Results of average accuracy for each classifier and dataset as a base classifier in Dagging}
\label{tab8}
\centering
\resizebox{1.05\textwidth}{!}{
\begin{tabular}{llc|llllllllllll}
\hline
Variability& dataset&Window size& DT& C-DT& RT& DS& NB& SVM& k-NN  & JRip  & OneR & CR& BN& DTNB\\\hline\hline
\multirow{18}{*}{Low}  & \multirow{6}{*}{Helpdesk}         &
3 & 65.98 & 65.23 & 65.49 & 62.76 & 62.63 & 41.71 & 65.36 & 64.04 & 64.68 & 59.22 & 63.06 & 64.90 \\
&&4 & 73.51 & 73.62 & 72.54 & 72.09 & 70.19 & 47.97 & 72.35 & 72.27 & 71.98 & 71.22 & 70.27 & 72.09 \\
&&5 & 75.16 & 73.17 & 72.69 & 72.33 & 71.43 & 43.35 & 71.79 & 72.45 & 73.29 & 71.30 & 70.16 & 72.21 \\
&&6 & 77.20 & 71.50 & 74.23 & 73.32 & 72.81 & 48.92 & 75.53 & 74.22 & 74.10 & 68.64 & 71.12 & 74.10 \\
&&7 & 67.70 & 56.54 & 67.98 & 66.55 & 69.14 & 45.49 & 70.91 & 61.52 & 64.76 & 55.65 & 66.52 & 70.29 \\
&&8 & 48.73 & 51.87 & 63.41 & 62.44 & 53.34 & 43.28 & 68.63 & 51.87 & 55.50 & 49.65 & 66.55 & 51.81 \\
\cline{2-15}

& \multirow{6}{*}{Hospital Billing} &
3 & 92.49 & 92.38 & 92.32 & 61.43 & 90.82 & 67.16 & 92.24 & 92.31 & 92.27 & 61.43 & 91.08 & 92.24 \\
&&4 & 91.10 & 90.69 & 91.00 & 73.42 & 87.11 & 66.49 & 90.55 & 90.22 & 90.55 & 73.42 & 87.38 & 90.70 \\
&&5 & 88.13 & 87.68 & 87.50 & 58.26 & 82.79 & 54.49 & 86.82 & 86.56 & 87.40 & 58.26 & 83.51 & 87.14 \\
&&6 & 85.51 & 84.73 & 84.64 & 44.23 & 78.39 & 31.57 & 82.86 & 83.52 & 83.60 & 44.18 & 78.90 & 82.94 \\
&&7 & 84.89 & 84.22 & 84.57 & 48.82 & 76.16 & 31.60 & 81.92 & 82.83 & 84.03 & 48.77 & 76.53 & 82.48 \\
&&8 & 87.56 & 86.58 & 87.35 & 54.77 & 76.68 & 34.02 & 84.11 & 84.76 & 86.27 & 54.74 & 78.58 & 85.78 \\
\cline{2-15}

& \multirow{6}{*}{Road Traffic}     &
3 & 94.65 & 92.00 & 95.65 & 77.64 & 88.49 & 41.55 & 95.49 & 93.84 & 85.26 & 77.62 & 89.63 & 91.73 \\
&&4 & 87.15 & 86.54 & 88.31 & 79.04 & 84.70 & 23.67 & 87.78 & 87.42 & 81.37 & 79.11 & 85.01 & 85.75 \\
&&5 & 83.77 & 81.86 & 87.92 & 61.07 & 71.69 & 26.16 & 85.31 & 81.98 & 74.80 & 61.87 & 68.78 & 77.65 \\
&&6 & 75.33 & 67.59 & 78.47 & 66.93 & 76.60 & 26.82 & 70.67 & 65.38 & 67.27 & 54.80 & 69.79 & 67.45 \\
&&7 & 75.12 & 48.14 & 72.73 & 65.40 & 81.51 & 34.30 & 70.33 & 44.68 & 65.40 & 54.29 & 74.50 & 73.68 \\
&&8 & 64.93 & 48.82 & 67.68 & 71.40 & 76.91 & 45.81 & 68.24 & 48.82 & 55.33 & 46.32 & 71.29 & 68.31
\\
\hline
\multicolumn{3}{l|}{Average Friedman rank} &\textbf{2.72}& 	\textit{5.06}& 	\textbf{2.72} &	8.42 &	7.17 &	11.89 &	\underline{4.39} &	6.28 &	6.08& 	10.53& 	7.22 &	5.53 
 \\\hline

\multirow{18}{*}{High}           & \multirow{6}{*}{BPIC2012}          &
3 & 75.45 & 75.48 & 74.65 & 73.44 & 67.33 & 15.27 & 74.22 & 75.23 & 73.72 & 73.44 & 67.54 & 75.38 \\
&&4 & 76.47 & 76.48 & 75.82 & 76.15 & 76.12 & 38.74 & 75.39 & 76.22 & 76.19 & 76.13 & 76.44 & 76.18 \\
&&5 & 71.20 & 71.11 & 69.33 & 71.06 & 71.20 & 57.17 & 68.72 & 70.73 & 71.37 & 71.03 & 71.20 & 71.26 \\
&&6 & 69.58 & 63.84 & 67.57 & 68.19 & 69.49 & 57.92 & 67.11 & 68.59 & 69.67 & 66.83 & 69.21 & 70.50 \\
&&7 & 64.97 & 63.51 & 65.51 & 64.37 & 64.17 & 49.66 & 65.58 & 64.44 & 65.30 & 63.51 & 63.57 & 65.37 \\
&&8 & 67.51 & 67.51 & 66.51 & 66.33 & 63.06 & 60.30 & 67.09 & 67.67 & 62.61 & 67.51 & 63.05 & 69.76 \\

\cline{2-15}

& \multirow{5}{*}{BPIC2013 Incident} &
3 & 64.67 & 64.37 & 63.00 & 50.17 & 60.50 & 41.34 & 62.93 & 51.99 & 61.65 & 51.31 & 62.71 & 63.92 \\
&&4 & 61.74 & 61.40 & 60.14 & 52.19 & 57.44 & 36.28 & 59.80 & 48.04 & 58.38 & 53.39 & 58.83 & 60.78 \\
&&5 & 63.70 & 63.32 & 62.55 & 49.20 & 58.68 & 33.26 & 61.98 & 50.89 & 60.21 & 49.20 & 60.88 & 62.68 \\
&&6 & 62.27 & 62.19 & 61.30 & 47.30 & 57.77 & 28.08 & 60.49 & 49.12 & 58.94 & 47.30 & 59.57 & 61.58 \\
&&7 & 62.75 & 62.83 & 61.55 & 48.66 & 58.36 & 22.43 & 60.72 & 50.71 & 59.77 & 48.66 & 59.80 & 62.18 \\
&&8 & 62.41 & 62.31 & 60.96 & 53.27 & 57.65 & 16.19 & 59.52 & 50.59 & 59.16 & 53.21 & 59.13 & 61.64 \\
\cline{2-15}

& \multirow{5}{*}{Sepsis}           &
3 & 53.06 & 51.74 & 52.12 & 42.02 & 48.21 & 28.48 & 51.64 & 49.87 & 48.66 & 41.4  & 49.07 & 50.6  \\
&&4 & 56.74 & 53.30 & 57.58 & 43.35 & 51.97 & 26.63 & 57.65 & 54.86 & 49.84 & 42.34 & 52.64 & 54.06 \\
&&5 & 57.11 & 53.69 & 58.19 & 37.09 & 51.46 & 23.96 & 57.37 & 54.82 & 49.38 & 40.37 & 52.27 & 54.48 \\
&&6 & 55.60 & 53.24 & 56.17 & 38.12 & 53.30 & 23.66 & 54.36 & 55.15 & 51.15 & 40.34 & 53.63 & 55.27 \\
&&7 & 55.30 & 54.18 & 54.93 & 34.73 & 52.86 & 22.73 & 51.99 & 55.29 & 53.53 & 35.15 & 53.01 & 55.06 \\
&&8 & 54.11 & 53.39 & 52.19 & 42.74 & 52.38 & 26.11 & 47.98 & 53.65 & 53.25 & 45.68 & 52.22 & 53.63
\\
\hline
\multicolumn{3}{l|}{Average Friedman rank} &\textbf{2.11}& 	\textit{4.25} &	4.67 &	9.44 &	7.72 &	12.00& 	5.78 &	6.17 &	6.50 &	9.53&	6.61& 	\underline{3.22} 
\\\hline
\end{tabular}
}
\end{table}
\end{appendices}

\end{document}